\documentclass{article} % For LaTeX2e
\usepackage{iclr2027_conference,times}
\usepackage{enumitem}
\usepackage{graphicx}
\usepackage{booktabs}
\usepackage{wrapfig}
\usepackage{amsmath,amsfonts,bm}

\def\eqref#1{equation~\ref{#1}}
\def\1{\bm{1}}

\DeclareMathAlphabet{\mathsfit}{\encodingdefault}{\sfdefault}{m}{sl}
\SetMathAlphabet{\mathsfit}{bold}{\encodingdefault}{\sfdefault}{bx}{n}

\usepackage{hyperref}
\usepackage{url}

\title{Retargeting Motions to Diverse Skeletons via Learnable Flattening}

\author{
Kia-Jüng Yang$^{1}$ \quad Fabian H. Sinz$^{1,2}$ \quad Paweł A. Pierzchlewicz$^{1,3}$
\\[0.5em]
{\small $^{1}$Institute of Computer Science, University of Göttingen}
\\
{\small $^{2}$Campus Institute Data Science, University Göttingen}
\\
{\small $^{3}$Pantomim P.S.A}
\\[0.3em]
{\small \texttt{kia-jueng.yang@uni-goettingen.de} \quad \texttt{sinz@cs.uni-goettingen.de} \quad \texttt{paul@animatica.ai}}
}
\iclrfinalcopy % Uncomment for camera-ready version, but NOT for submission.
\begin{document}

\maketitle

\begin{abstract}

Cross-structural motion retargeting aims to transfer motion between different skeletal topologies. Despite recent progress, existing state-of-the-art models struggle with reliability in zero-shot settings, i.e. skeletons with different topologies which were unseen during training, and recent Transformer-based attempts have failed to outperform specialized geometric methods. We bridge this gap with a Transformer Autoencoder that learns a topology- and translation-invariant latent space. Our core contribution is a learnable flattening of skeletal graphs that captures both local dependencies and global structure. Unlike the standard transformer architecture, which adds positional information to token content, we integrate graph-based positional encodings multiplicatively, a design choice that follows directly from our flattening formulation. The resulting model handles diverse skeletal topologies within a single unified architecture and trains in a fully unsupervised manner, requiring no paired retargeting data. Ablation studies show, that the graph encodings, multiplicative formulation, and Transformer backbone is critical for the performance. In zero-shot evaluations, our method reduces global joint position error by $43-47\%$ over current benchmarks. A user study ($n = 37$), including expert animators, further ranks our approach highest in motion alignment and physical plausibility ($p < 0.05$). These results demonstrate that our model design is key to making transformer architectures effective for motion retargeting, outperforming existing approaches.
\end{abstract}

\section{INTRODUCTION}

Character animation and motion synthesis are longstanding challenges in computer graphics and computer vision, with broad applications in entertainment, virtual reality, biomechanics, and human-computer interaction~\citep{bruderlin1995motion,Gleicher1998,holden2016deep,loper2015smpl}. 
Among these, particularly motion retargeting---the process of transferring motion between characters with differing skeletal topologies---is challenging. Traditional approaches often rely on manual efforts by skilled animators or require specialized algorithms tailored for specific pairs of skeletons, limiting scalability and generalization.

Deep learning approaches have recently shown promising results in motion synthesis and representation~\citep{pavllo20193dhpe,petrovich2021action,tevet2022human,yuan2022physdiff}. However, most existing methods operate on fixed skeletal topologies, limiting their generalizability across diverse character structures. The underlying challenge lies in the representation of skeletal motion data, which inherently combines local joint rotations with global hierarchical structure. When mapped to a latent space, these representations often become tied to specific skeletal configurations, making cross-skeleton transfer difficult.

\begin{figure}[t]
    \centering
    \includegraphics[width=0.9\linewidth]{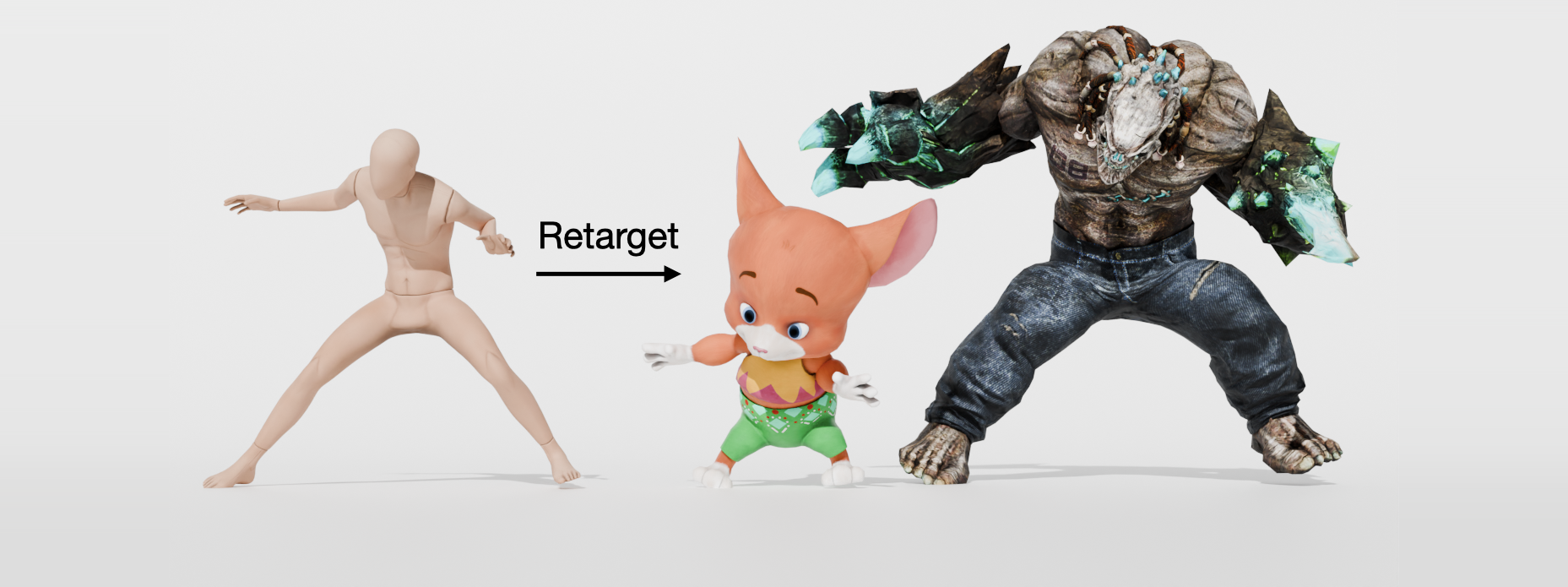}
    \caption{Our method retargets a source motion (left) to characters with vastly different skeletal topologies via learnable flattening of skeletal graphs in a fully zero-shot setting.}
    \label{fig:teaser}
\end{figure}

In this paper, we introduce a Transformer Autoencoder for motion retargeting that learns a skeleton topology- and translation-invariant latent space, enabling seamless motion transfer across characters with vastly different skeletal structures (Figure~\ref{fig:teaser}). The central architectural contribution is a learnable flattening of skeletal graphs, which reinterprets the standard joint-concatenation operation as a combination of learnable tiling and masking. This formulation allows the model to generalize to unseen skeletal topologies and directly motivates our integration of graph-based positional encodings via a multiplicative scheme, a design choice our ablations confirm is critical for retargeting quality.

The model is trained end-to-end in a fully unsupervised, cycle-consistent manner, requiring no paired retargeting data, and incorporates novel data augmentation strategies that our ablations confirm are critical for zero-shot generalization. The resulting framework outperforms all existing methods both quantitatively and qualitatively, substantially advancing the state of the art in motion retargeting across unseen skeletal topologies.
Our main contributions are:

\begin{itemize}[leftmargin=*]
    \item A Transformer Autoencoder that retargets motion between arbitrary humanoid skeletons — with varying numbers of joints and edges — within a single unified model, trained without paired data, or textual joint information, encoding skeletal structure purely from graph topology and rest pose geometry.
    \item A learnable skeletal flattening with multiplicative graph-positional encodings that yields a topology- and translation-invariant latent space, directly enabling zero-shot generalization to unseen skeletons.
    \item State-of-the-art retargeting performance, reducing global joint position error by 43–47\% over existing benchmarks and ranking highest in motion alignment and physical plausibility in a user study with expert animators ($n=37$, $p<0.05$).
\end{itemize}

\section{RELATED WORK}

Our work builds upon and extends research in several areas, including character animation, motion retargeting and deep representation learning for skeletal motion. We review the most relevant contributions in each of these domains.

\subsection{Motion Retargeting}

Motion retargeting transfers motion between characters while preserving semantic meaning. Classical approaches formulate this as space-time optimization~\citep{Gleicher1998, Tak2005} or inverse kinematics~\citep{Lee1999, Choi2000}, requiring extensive manual tuning. Recent deep learning methods are largely unsupervised due to the lack of paired datasets. 

Villegas et al.~\citeyearpar{Villegas2018} use an RNN with cycle consistency and adversarial training, but perform poorly across different skeleton structures. Aberman et al.~\citeyearpar{Aberman2020} support arbitrary joint counts but need a separate model per skeleton pair. Lee et al.~\citeyearpar{Lee2023SAME} take a supervised approach, relying on MotionBuilder~\citep{autodesk_motionbuilder} to generate proxy ground truth. Liu et al.~\citeyearpar{liu2026palum} propose a transformer-based unsupervised single model but rely on textual joint name embeddings and still fail to outperform existing methods, indicating that architecture alone is insufficient without careful design. Chen et al.~\citeyearpar{chen2025motion2motion} offer a training-free alternative via patch-based motion matching, but require explicit bone correspondences and a target motion database at inference time.

Our work is the first to make transformers effective for motion retargeting via a learnable flattening of skeletal graphs, outperforming all existing methods with a single unsupervised zero-shot model.

\subsection{Deep Motion Representations}

Learning compact latent motion representations has proven highly effective across tasks. Holden et al.~\citeyearpar{holden2015learning} pioneered convolutional autoencoders for motion, Aberman et al.~\citeyearpar{aberman2019learning} learned a skeleton-agnostic latent space for 2D retargeting, Athanasiou et al.~\citeyearpar{athanasiou2022teach} encode spatio-temporal motion with text for conditional generation, and Starke et al.~\citeyearpar{starke2022deepphase} capture motion in periodic feature embeddings.

MotionPuzzle~\citeyearpar{jang2022motionpuzzle} employs cycle-consistency and reconstruction objectives on unpaired data for per-body-part style transfer, but within a fixed topology. Gat et al.~\citeyearpar{gat2025anytop} integrate graph structure additively into transformer attention maps. In contrast, our multiplicative positional encoding is derived from first principles and our ablations confirm it is critical for zero-shot generalization.

Despite large-scale successes in NLP and vision, motion data remains fragmented across incompatible parametrizations. AMASS~\citep{mahmood2019amass} unified datasets into SMPL~\citep{loper2015smpl,pavlakos2019expressive}, and learned motion priors~\citep{rempe2021humor,chen2022learning,raab2023single,yuan2022physdiff} offer unified representations, but all remain limited to fixed humanoid topologies. SAME~\citeyearpar{Lee2023SAME} handles varying topologies in a single model but requires ground-truth retargeting pairs. 

In contrast, our framework requires neither, and learns a skeleton-agnostic pose component alongside a separate root trajectory component.

\section{METHODS}\label{sec:methods}

% To enable motion retargeting across diverse skeletal topologies, we must decouple the motion representation from the specific joint count and structure of the source character. Traditional approaches flatten joint features into high-dimensional vectors (e.g., $\mathbb{R}^{J \times d} \rightarrow \mathbb{R}^{Jd}$), creating representations that scale linearly with the number of joints $J$. This structural coupling prevents generalization to new skeletons. We address this by formally decomposing the flattening operation, revealing that it is implicitly a \textit{multiplicative positional encoding} process. By replacing the fixed operations of standard flattening with learned neural networks, we obtain fixed-dimensional latent representations that preserve spatial information while remaining topology-agnostic.

In order to enable retargeting across diverse skeletal topologies, we must decouple the motion representation 
from the source skeleton's joint count and structure. Standard flattening 
($\mathbb{R}^{J \times d} \rightarrow \mathbb{R}^{Jd}$) creates representations 
that scale linearly with $J$, preventing generalization to new skeletons. We address 
this by decomposing the flattening operation, revealing it is implicitly a 
\textit{multiplicative positional encoding} process, and replacing its fixed 
operations with learned neural networks to obtain fixed-dimensional, 
topology-agnostic latent representations.

\subsection{Data Representation} We first formalize the skeletal motion inputs used in our framework. The skeleton is defined by its rest pose $\mathcal{S} = \{p_{\text{rest}}, \mathcal{A}\}$. Here, $p_{\text{rest}} \in \mathbb{R}^{J \times 3}$ encodes the rest pose geometry based on fixed bone offsets in a canonical T-pose configuration (with the $y$-axis defined as up), while $\mathcal{A} \in \{0,1\}^{J \times J}$ is the adjacency matrix defining kinematic connectivity. The motion is modeled as a temporal sequence $\mathbf{X} = \{ q_t, p_t, p_{t-1}, v_t, r_t \}_{t=1}^{T}$. Here, $q_t \in \mathbb{R}^{J \times 6}$ encodes joint rotations using the continuous 6D representation \citep{zhou2019continuity}, $p_t \in \mathbb{R}^{J \times 3}$ denotes root-centered joint positions, $v_t$ represents joint velocities, and $r_t \in \mathbb{R}^3$ specifies the global root trajectory. %This separation allows us to process local joint dynamics ($q_t, p_t$) independently of global translation ($r_t$).

\subsection{Flattening Representation Learning}

\paragraph{Decomposing Flattening.} Consider a pose on a skeleton with $J$ joints, each represented by $d$ features, forming a data matrix $\mathcal{X} \in \mathbb{R}^{J \times d}$. Standard flattening concatenates the rows of $\mathcal{X}$ into a single vector $\bar{q} = \text{vec}(\mathcal{X}^T) \in \mathbb{R}^{Jd}$. We can express this operation as the interaction between two matrices: a projection matrix $\mathbf{T}$ and a positional mask $\mathbf{M}$.
\begin{equation}
\mathbf{T} = \bigl[ \mathbf{I}_d,\, \mathbf{I}_d,\, \ldots,\, \mathbf{I}_d \bigr],\quad \mathbf{M} = \mathbf{I}_J \otimes \mathbf{1}_d^T \quad
\end{equation}
The projection matrix $\mathbf{T} \in \mathbb{R}^{d \times Jd}$ broadcasts the content features of a joint across all possible output positions and $\otimes$ denotes the Kronecker product. The positional mask $\mathbf{M} \in \mathbb{R}^{J \times Jd}$ is a sparse binary matrix that "activates" the specific slot in the flattened vector corresponding to the $j$-th joint. The flattening and unflattening operations can then be rewritten as:
\begin{align}
\bar{q} &= \sum_{j=1}^{J} \Bigl[ (\mathcal{X}_j \mathbf{T}) \odot \mathbf{M}_j \Bigr], \label{eq:flatten_decomposition} \\
\mathcal{X} &= ((\mathbf{1}_J\bar{q}) \odot \mathbf{M}) \mathbf{T}^T, \label{eq:unflatten_decomposition}
\end{align}
where $\odot$ denotes the Hadamard (element-wise) product and the subscript $j$ denotes the $j$-th row of the corresponding matrix.

The (un)flattening operation functions as a \textit{structural gating mechanism}, mathematically analogous to attention, where a map $\mathbf{A}$ applies to values $\mathcal{V}$. In our case, $\mathbf{M}_j$ acts as a static attention map $\mathbf{A}$ applied \textit{multiplicatively} to the values $\mathcal{V}$ ($\mathbf{X}_j \mathbf{T}$ or $\bar{q}$).
Since $\mathbf{M}_j$ encodes joint identity via the position of its nonzero entries, this acts as a positional encoder.
% From this perspective, (un)flattening functions as a gating mechanism where $\mathbf{A}$ is applied \textit{multiplicatively} to $\mathcal{V}$.$\mathbf{M}_j$ encodes the joint identity: the position of its nonzero entries acts as a positional encoder. This formulation highlights that
Unlike standard positional encoding which typically sum position and content, here the content ($\mathcal{X}_j \mathbf{T}$) is modulated by the structural position ($\mathbf{M}_j$) \textit{via multiplication}.

\paragraph{Learnable Flattening.} To achieve a topology-agnostic representation, we replace these fixed matrices with learned, continuous functions. Our encoder implements the learnable flattening operation:
\begin{equation}
z_p = \mathsf{Agg}_{j\in J} \Bigl[\mathsf{T}_{\text{enc}}(\mathcal{X}_j) \odot \mathsf{M}(\mathcal{S})_j \Bigr], \label{eq:encoder} \quad \text{}
\end{equation}
where $\mathsf{T}_{\text{enc}}: \mathbb{R}^d \rightarrow \mathbb{R}^D$ is a learned projection that maps joint features to a fixed latent dimension $D$, independent of $J$. $\mathsf{M}(\mathcal{S}) \in \mathbb{R}^{J \times D}$ is a learned positional embedding derived dynamically from the skeleton topology $\mathcal{S}$.

This formulation preserves the multiplicative interaction crucial for spatial structure but decouples the representation size from the joint count. The decoder reverses this operation (``unflattening'') to reconstruct the specific skeletal motion:
\begin{equation}
\hat{\mathcal{X}} = \mathsf{T}_{\text{dec}}\Bigl( z_p \odot \mathsf{M}(\mathcal{S}) \Bigr), \label{eq:decoder}
\end{equation}
where $\mathsf{T}_{\text{dec}}: \mathbb{R}^D \rightarrow \mathbb{R}^d$ reconstructs per-joint features from the latent space.

\subsection{Model Architecture}
\begin{figure}[t]
  \centering
  \includegraphics[width=0.6\linewidth]{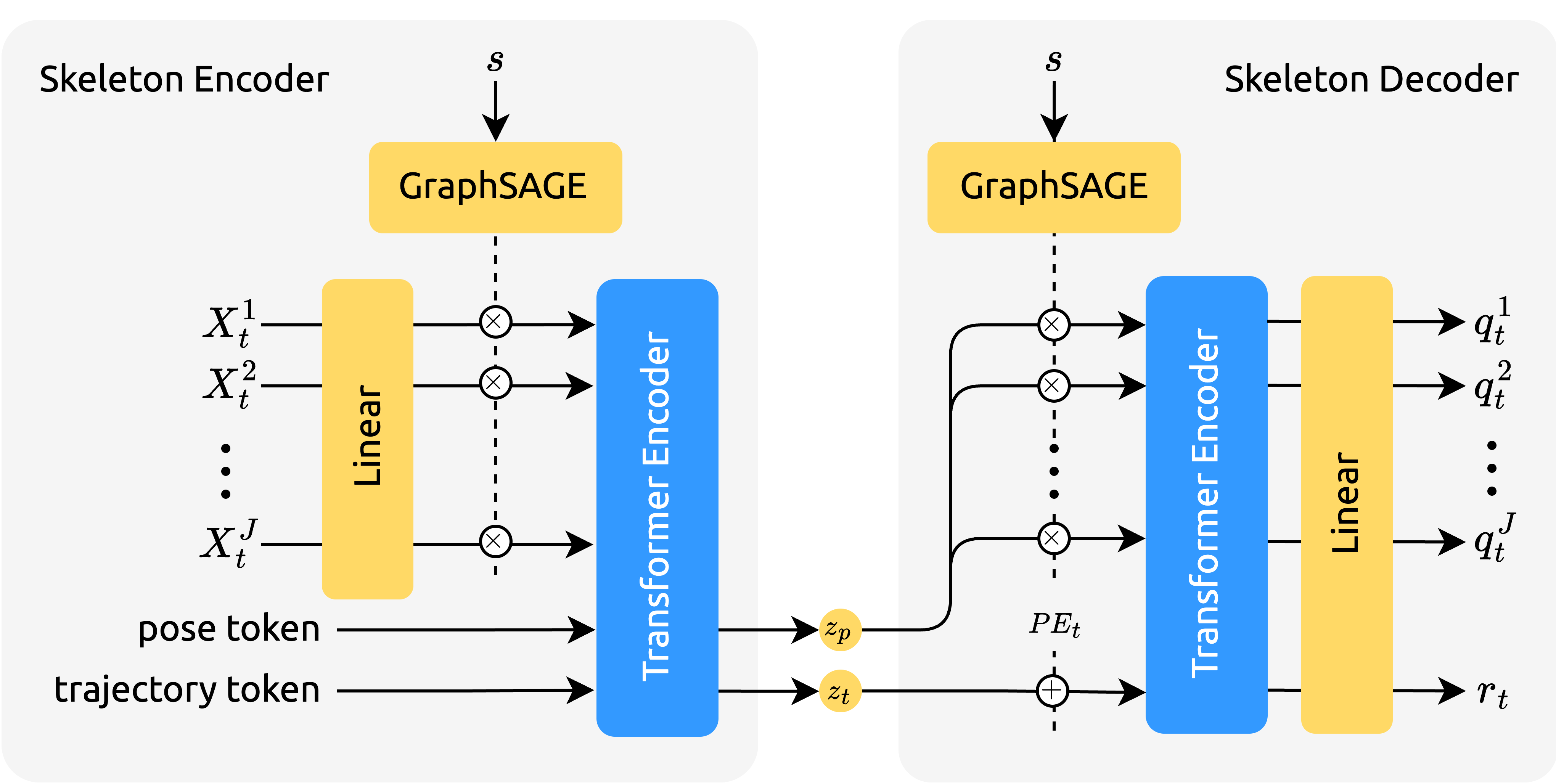}
  \caption{%Schematic of the SkIP autoencoder. The Skeleton Position Encoding module uses GraphSAGE to produce a spatial mask from the rest pose $\mathcal{S}$. The Skeleton Encoder "flattens" skeletal data into latent pose and trajectory tokens via a transformer encoder, while the Skeleton Decoder "unflattens" those tokens to recover joint rotations.
  Schematic of the our model architecture. The Skeleton Position Encoding uses GraphSAGE to produce a spatial mask from the rest pose $\mathcal{S}$. The Encoder "flattens" skeletal data into latent pose and trajectory tokens, while the Decoder "unflattens" them to recover joint rotations.}
  \label{fig:architecture}
\end{figure}

% Our framework leverages a transformer-based autoencoder that implements the learned flattening and unflattening operations. It integrates three key modules: (1) a \textbf{Skeleton Position Encoding}, which learns the spatial mask $\mathbf{M}(\mathcal{S})$; (2) a \textbf{Skeleton Encoder}, which performs the learnable flattening (Eq.~\ref{eq:encoder}) to produce latent tokens; and (3) a \textbf{Skeleton Decoder}, which performs the unflattening (Eq.~\ref{eq:decoder}) to reconstruct the motion. Figure~\ref{fig:architecture} illustrates this pipeline. For architecture details please refer to the Appendix Section \ref{sec:architecture_details}.
Our framework leverages a transformer-based autoencoder that implements the learned flattening and unflattening operations. It integrates three key modules: (1) a \textbf{Skeleton Position Encoding}, which learns the spatial mask $\mathbf{M}(\mathcal{S})$; (2) a \textbf{Skeleton Encoder}, which performs the learnable flattening (Eq.~\ref{eq:encoder}) to produce latent tokens; and (3) a \textbf{Skeleton Decoder}, which performs the unflattening (Eq.~\ref{eq:decoder}) to reconstruct the motion (Figure~\ref{fig:architecture}). For architecture details please refer to the Appendix Section \ref{sec:architecture_details}.

\paragraph{Skeleton Position Encoding.}
$\mathsf{M}: \mathcal{S} \rightarrow \mathbb{R}^{J \times D}$ generates learned spatial masks from the rest pose $\mathcal{S}$ using an undirected GraphSAGE~\citep{hamilton2017inductive} network, producing a spatially-aware embedding $\mathsf{M}(\mathcal{S})_j$ for topology-aware encoding and decoding.

\paragraph{Skeleton Encoder.} Implements the \textit{learnable flattening} in Equation~\ref{eq:encoder}. First a linear projection layer $\mathsf{T}_{\text{enc}}: \mathbb{R}^d \rightarrow \mathbb{R}^D$ acts as the projection operator, mapping each joint's features into a shared $D$-dimensional space, where it is element-wise multiplied by the learned spatial mask $\mathsf{M}(\mathcal{S})$. A Transformer $\mathsf{Agg}$ then aggregates the joint features into a fixed-dimensional pose token $z_p \in \mathbb{R}^D$, while a separate trajectory token $z_t$ encodes global translation independently of pose.

\paragraph{Skeleton Decoder.} Implements the \textit{learnable unflattening} in Equation~\ref{eq:decoder}. The spatial mask $\mathsf{M}(\mathcal{S})$ is reapplied to the pose token $z_p$, and a Transformer decoder combines the resulting features with the trajectory token $z_t$. Finally, a linear layer maps the decoded features back to the original input dimensions: 6 for joint rotations and 3 for the root trajectory. Together, the Transformer and linear layer constitute the learned decoding operator $\mathsf{T}_{\text{dec}}$. We use forward kinematics to obtain world-space joint positions.

\subsection{Training}
\begin{wrapfigure}{r}{0.55\linewidth}
    \centering
    \includegraphics[width=\linewidth]{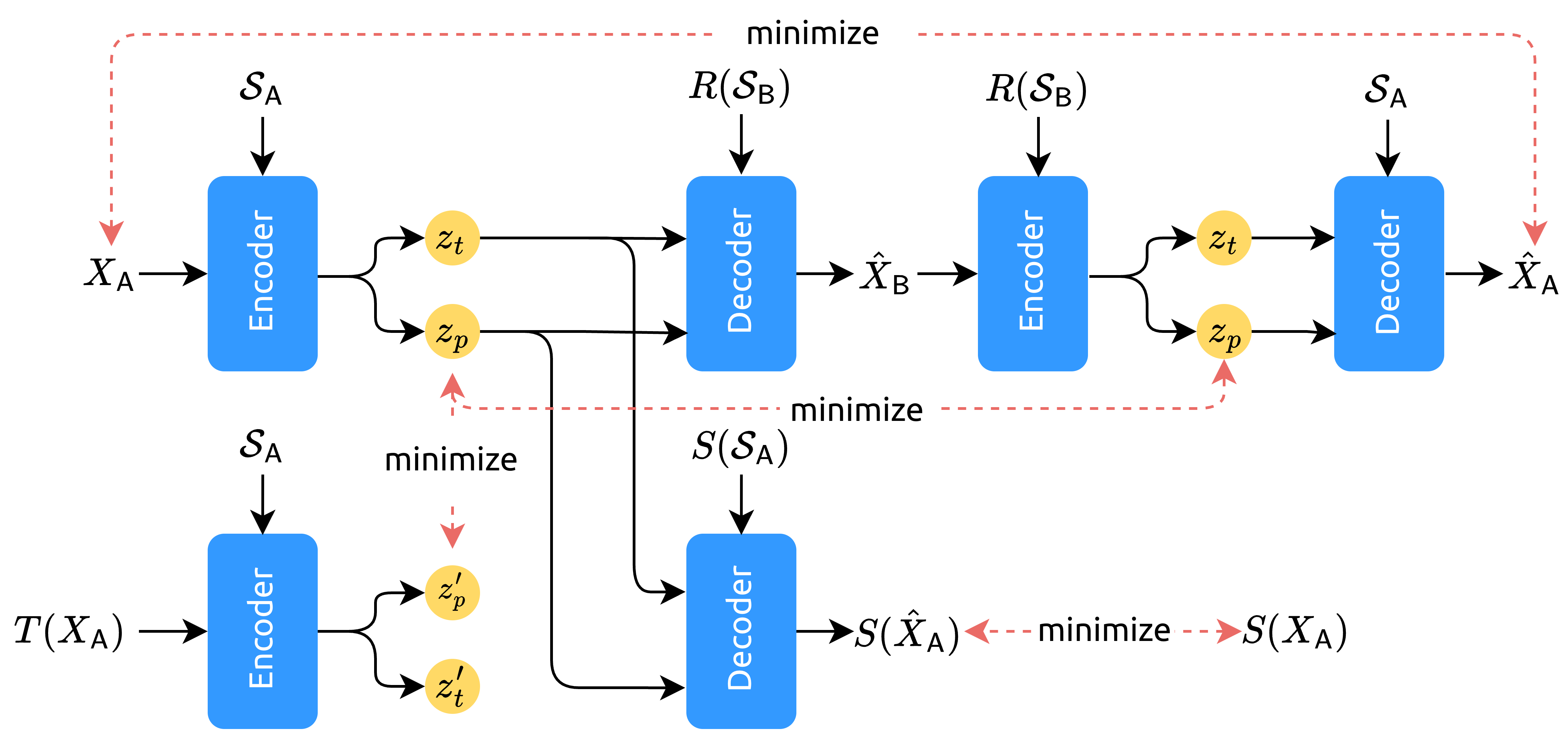}
    \caption{Schematic of the training framework.}
    \label{fig:training}
\end{wrapfigure}

In the absence of explicit ground truth, our framework uses cycle 
consistency~\citep{zhu2017unpaired}, an adversarial loss~\citep{goodfellow2014generative}, 
and an augmentation consistency strategy (Figure~\ref{fig:training}). Three augmentations 
prime the autoencoder to learn meaningful invariances and execute reliable retargeting: global translation $T(\cdot)$, 
bone scaling $S(\cdot)$, and skeleton augmentation $R(\cdot)$ (per-joint scaling and joint removal).

The losses are defined as follows: $\mathcal{L}_\text{rec}$ combines joint-position, 
rotation, root-trajectory, and root-child position losses (the latter introduced to 
preserve joints directly connected to the root). $\mathcal{L}_\text{temp}$ encourages 
temporal smoothness, $\mathcal{L}_\text{cntct}$ reduces foot sliding and ground 
penetration, $\mathcal{L}_\text{ee}$ preserves normalized end-effector velocities, 
$\mathcal{L}_\text{latent}$ enforces pose consistency, and $\mathcal{L}_\text{adv}$ 
encourages realistic motions. Except for the root-child position loss, these follow 
prior work~\citep{Aberman2020,Lee2023SAME}. Details are provided in 
Appendix~\ref{app_sec:loss_func_weighting}.

Two complementary schemes combine these losses: \textit{augmentation consistency} and \textit{cycle consistency}. Given an input motion $\mathcal{X}_\mathsf{A}$ associated with a source skeleton $\mathcal{S}_\mathsf{A}$.
Denote the encoder and decoder by $\mathcal{E}$ and $\mathcal{D}$, respectively.
The augmented reconstruction $S(\hat{\mathcal{X}}_\mathsf{A})$ is computed as encoding the input motion $\mathcal{X}_\mathsf{A}$ with skeleton $\mathcal{S}_\mathsf{A}$ and then decoding into the augmented skeleton $S(\mathcal{S}_\mathsf{A})$.
Cycle consistency is enforced by retargeting the input motion $\mathcal{X}_\mathsf{A}$ with skeleton $\mathcal{S}_\mathsf{A}$ to a target skeleton $R(\mathcal{S}_\mathsf{B})$ obtaining the retargeted $R(\hat{\mathcal{X}}_\mathsf{B})$ and then mapping it back to the original skeleton $\mathcal{S}_\mathsf{A}$ obtaining $\hat{\mathcal{X}}_\mathsf{A}$. For $A \in \{S, T, R\}$, let 
$\mathcal{X}^A = A(\mathcal{X})$:

\begin{align*}
    \mathcal{L}_\text{aug. cons.}
    &= \mathcal{L}_\text{rec}(\mathcal{X}^S_\mathsf{A}, \hat{\mathcal{X}}^S_\mathsf{A})
    + \mathcal{L}_\text{cntct}(\mathcal{X}^S_\mathsf{A}, \hat{\mathcal{X}}^S_\mathsf{A})
    + \mathcal{L}_\text{temp}(\mathcal{X}^S_\mathsf{A}, \hat{\mathcal{X}}^S_\mathsf{A})
    + \mathcal{L}_\text{ee}(\mathcal{X}_\mathsf{A}, \hat{\mathcal{X}}^S_\mathsf{A})
    + \mathcal{L}_\text{latent}(\mathcal{X}_\mathsf{A}, \mathcal{X}^T_\mathsf{A}) \\
    \mathcal{L}_\text{cyc. cons.}
    &= \mathcal{L}_\text{rec}(\mathcal{X}_\mathsf{A}, \hat{\mathcal{X}}_\mathsf{A})
    + \mathcal{L}_\text{cntct}(\mathcal{X}_\mathsf{A}, \hat{\mathcal{X}}_\mathsf{A})
    + \mathcal{L}_\text{latent}(\mathcal{X}_\mathsf{A}, \hat{\mathcal{X}}^R_\mathsf{B})
    + \mathcal{L}_\text{adv}(\mathcal{X}_\mathsf{A}, \hat{\mathcal{X}}^R_\mathsf{B})
\end{align*}

The total loss is $\mathcal{L}_\text{aug. cons.} + \mathcal{L}_\text{cyc. cons.}$.

\section{EXPERIMENTS}\label{sec:experiments}

We assess three capabilities: zero-shot reconstruction, latent space invariance, and unsupervised retargeting fidelity. We compare against SAME~\citep{Lee2023SAME}, which handles arbitrary topologies in a single model via a delta-based trajectory approach, \citet{Aberman2020}, the standard for unsupervised retargeting with topology-specific models. and PALUM~\citep{liu2026palum}, an unsupervised transformer-based approach that also handles all joint topologies in a single model.

We train on LaFAN1~\citep{harvey2020robust}, ACCAD~\citep{accad}, 
TotalCapture~\citep{trumble2017totalcapture}, PFNN~\citep{holden2017phase}, 
SFU~\citep{sfu2011dataset}, and CMU~\citep{cmu2003mocap}. We evaluate on 
Bandai-Namco~\citep{kobayashi2023motion} and 100Style~\citep{mason2022local} as 
zero-shot benchmarks, and Mixamo~\citep{adobe2021mixamo} for intra- and cross-structure 
retargeting comparisons. Further details are in Appendix~\ref{app_sec:data_prep}.

\subsection{Reconstruction}

\begin{wraptable}{r}{0.5\textwidth}
  \centering
  \vspace{-\intextsep}
  \caption{Reconstruction Results on unseen testing datasets. We compare Joint Position (JP), Joint Rotation (JR), Root Trajectory (RT), Foot Sliding (FS), and Ground Penetration (GP) errors. \textbf{Bold} indicates best.}
  \small
  \begin{tabular}{lccccc}
    \toprule
    \textbf{Method} & \textbf{JP} & \textbf{JR} & \textbf{RT} & \textbf{FS} & \textbf{GP} \\
    & [cm] & [rad] & [cm] & & [cm] \\
    \midrule
    \multicolumn{6}{l}{\textit{Bandai-Namco (Unseen)}} \\
    SAME & 12.40 & 0.34 & 8.58 & 0.08 & -0.20 \\
    Ours & \textbf{2.63} & \textbf{0.19} & \textbf{1.65} & \textbf{0.05} & \textbf{-0.01} \\
    \midrule
    \multicolumn{6}{l}{\textit{Mixamo (Unseen)}} \\
    SAME & 10.07 & 0.31 & 4.47 & \textbf{0.02} & \textbf{-0.02} \\
    Ours & \textbf{4.08} & \textbf{0.22} & \textbf{0.84} & 0.03 & \textbf{-0.02} \\
    \midrule
    \multicolumn{6}{l}{\textit{100 Style (Unseen)}} \\
    SAME & 104.66 & 0.22 & 101.78 & \textbf{0.01} & \textbf{-0.00} \\
    Ours & \textbf{2.14} & \textbf{0.16} & \textbf{1.85} & 0.02 & \textbf{-0.00} \\
    \bottomrule
  \end{tabular}
  \label{tab:evaluation_results}
\end{wraptable}

We evaluate zero-shot reconstruction on Mixamo, Bandai-Namco~\citep{kobayashi2023motion}, 
and 100Style~\citep{mason2022local} against SAME~\citep{Lee2023SAME}, the only prior 
work handling diverse unseen skeletons without retraining and publicly available code. We report Joint Position (JP) and Joint Rotation (JR) errors, Root Trajectory (RT) error, Foot Sliding (FS), and Ground Penetration (GP), following the evaluation protocol of \citet{Lee2023SAME}. Critically, we evaluate on 
\textit{entire animation sequences} instead of SAME's 1-second clips, to allow quantitative assessment of long-term stability.

Our method generalizes better to unseen datasets and outperforms SAME across all datasets (Table~\ref{tab:evaluation_results}). On Bandai-Namco, Joint Position error drops by 78\% 
(12.40$\rightarrow$2.63\,cm), with lower Foot Sliding (0.08$\rightarrow$0.05) and 
Ground Penetration (-0.20$\rightarrow$-0.01). Particularly, SAME requires ground contact labels at inference while we use them only during training. On 100Style, SAME suffers catastrophic drift (JP:104.66$\rightarrow$2.14\,cm, RT: 101.78$\rightarrow$1.85\,cm) due to delta-based root prediction accumulating errors over long sequences (Figure~\ref{fig:trajectory_analysis}, 
Appendix~\ref{appendix:traj_err_accumulation}). Unlike SAME, we predict the global 
root trajectory directly like~\citet{Aberman2020}, eliminating drift entirely.

\subsection{Latent Space Invariances}

\subsubsection{Skeleton Invariance}

\begin{figure*}[h]
  \centering
  \includegraphics[width=1\linewidth]{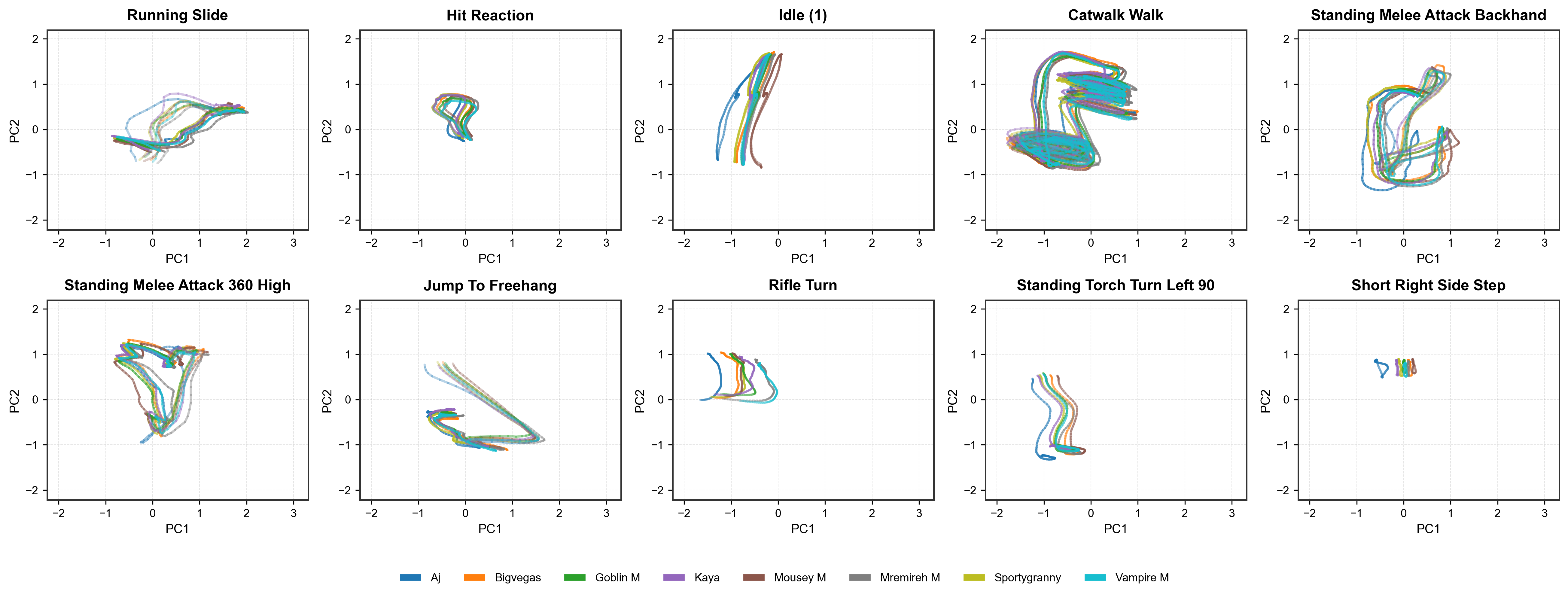}
  \caption{Skeleton invariance latent spaces of the model: 2D Principal Component Analysis projected latent space of the model for different skeletons performing semantically identical motions for different motions. The PCA space is shared across all sequences.}
  \label{fig:skeleton_invariance}
\end{figure*}

Our latent space is designed to be topology-agnostic, ensuring that identical motions map to the same representation regardless of the skeleton. To verify this, we project latent representations of distinct Mixamo characters (Mousey, Goblin, Vampire, etc.) into a 2D PCA space, following the approach of \citet{Lee2023SAME}. The tightly overlapping clusters show that embeddings from different skeletons (represented by different colors) align closely when performing the same action (e.g., "Walking," "Idle"), confirming that our model abstracts motion semantics while discarding skeleton-specific structure (Figure~\ref{fig:skeleton_invariance}).

\subsubsection{Translation Invariance}

\begin{figure*}[h]
  \centering
  \includegraphics[width=0.7\linewidth]{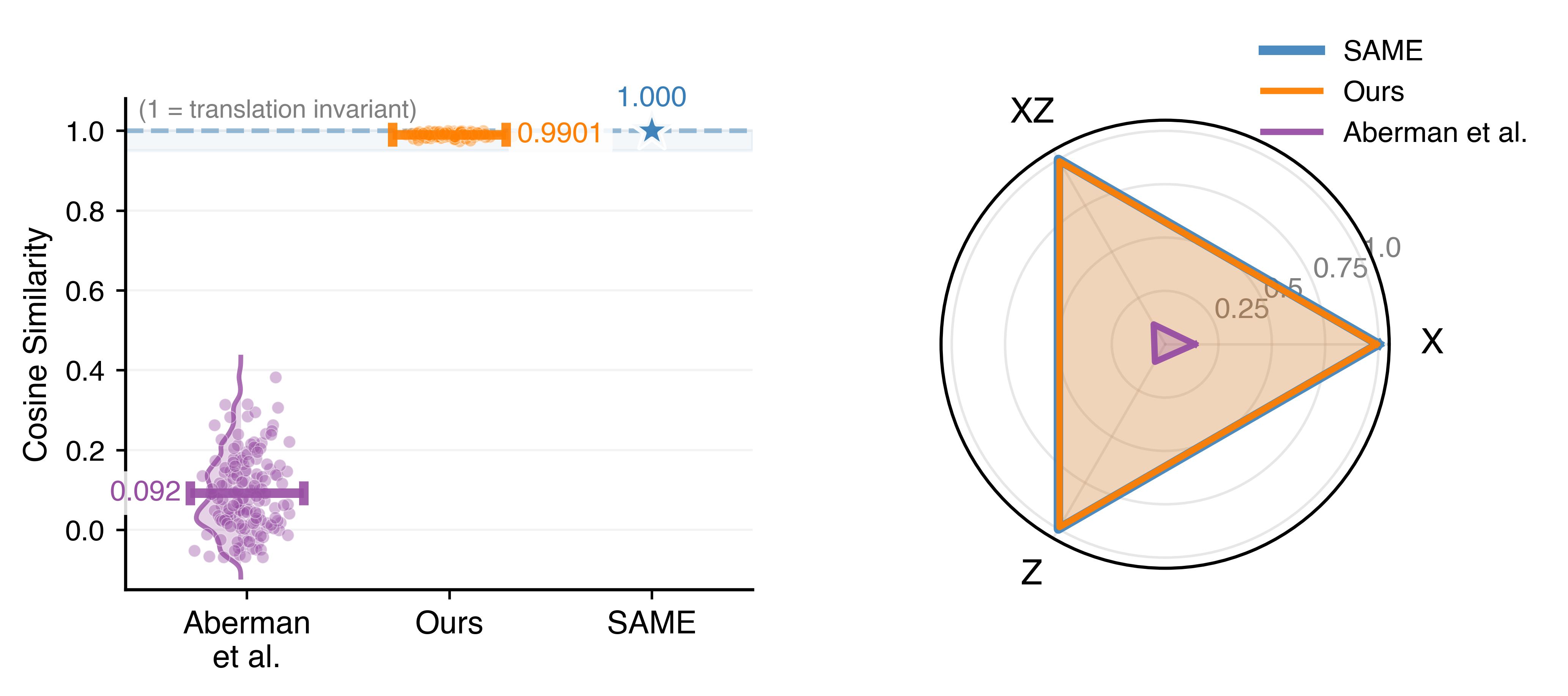}
  \caption{Translation invariance of our pose latent space. Left shows distributions of cosine similarities between latents of untranslated and translated motions. SAME is invariant by definition, so all the values are 1. Right shows the mean cosine similarities across translations in the x-, z- and xz-axes.}
  \label{fig:translation_invariance}
\end{figure*}

Our model disentangles motion into two distinct latent spaces --- pose ($z_p$) and root trajectory ($z_t$) --- with $z_p$ designed to be invariant to global translation. To validate this, we compare pose embeddings of identical motions before and after applying large-scale global translations ($10^6$ cm) along the $x$-, $z$-, and $xz$-axes, measuring Cosine Similarity between original and translated representations (Figure~\ref{fig:translation_invariance}). 

Our method achieves near-perfect invariance ($>0.99$), while SAME also reaches perfect invariance ($1.0$) as expected from its delta-based representation. In contrast, \citet{Aberman2020} shows significantly higher sensitivity, with scores dropping below $0.8$ and exhibiting high variance, confirming that it entangles global position with pose, a limitation our explicit disentanglement and augmentation strategy avoids.

\subsubsection{Motion Classification}

In order to measure that the semantic of the motion was captured across different skeletons, we ran a motion classification task based on our pose latent space. A topology-invariant latent space should cluster semantically similar motions together regardless of the skeletons they were performed on, and therefore the classification accuracy tells something about the quality of the latent space. Following the evaluation protocol of~\citet{Lee2023SAME}, we achieve 59/60 correctly classified motions compared to 57/60 reported by SAME. The only difference from their setup is that our latent space has 128 dimensions compared to 32 in SAME.

\subsection{Retargeting}
Motion retargeting transfers a motion sequence from one character to another with a different skeletal structure, while preserving the intent and style of the original movement. We evaluate our method on this task both qualitatively, through a user study, and quantitatively, using three complementary metrics across two structural settings.

\subsubsection{Quantitative Measurement}

To assess quantitative performance, we evaluate on two sub-tasks: \textit{Intra-structure} (Intra), where source and target share identical skeletal topologies, and the more challenging \textit{Cross-structure} (Cross), where topologies differ significantly. We use three metrics: \textbf{Global Joint Position (GJP)} error \citep{Aberman2020}, the mean Euclidean distance between predicted and ground-truth joint positions in global space, normalized by character height and scaled by $10^3$; \textbf{Jerk}, the third derivative of joint positions, where values closer to ground truth indicate natural dynamics; and \textbf{Procrustes Aligned Root Trajectory (PART)} error, which measures how well the predicted root trajectory preserves the original motion shape, independent of absolute alignment.

Despite not being trained on Mixamo, our method achieves state-of-the-art GJP across both Intra and Cross tasks, outperforming the next best method by at least 47\% and 43\% respectively (Table~\ref{tab:retargeting_results}). Our Jerk score (0.72) falls below ground truth ($\approx1.3$), reflecting a mild smoothing effect rather than a perceptual deficiency — adding minimal noise ($\sigma=0.005$) sufficient to produce visible jitter raises Jerk to 8.61, confirming the gap lies well below the threshold of visual perceptibility. PART scores further show that methods using global root representations (ours and \citet{Aberman2020}) preserve trajectory shape considerably better than delta-based methods (SAME).

\begin{table}[h]

\caption{%Animation Retargeting evaluation on the Mixamo dataset comparing Intra-structure and Cross-structure performance.
%\textbf{Bold} indicates best; \underline{underline} indicates second best.
%GT Jerk is 1.28 (Intra) and 1.32 (Cross).
%Note: ${\dagger}$ Method not trained on Mixamo motions. ${\ddagger}$ Method not trained on Mixamo skeletons.
Animation retargeting on Mixamo. \textbf{Bold}: best; \underline{underline}: second best. GT Jerk: 1.28 (Intra), 1.32 (Cross). 
${\dagger}$: not trained on Mixamo motions. 
${\ddagger}$: not trained on Mixamo skeletons.}
\centering
\begin{tabular}{lcccccc}
\toprule
& \multicolumn{3}{c}{Intra} & \multicolumn{3}{c}{Cross} \\
\cmidrule(lr){2-4} \cmidrule(lr){5-7}
Method & GJP & Jerk & PART & GJP & Jerk & PART \\
\midrule
Copy rotations & 8.86 & - & - & N/A & N/A & N/A\\
Villegas '18 & 6.24 & - & - & 243 & - & -\\
Lim '19 & 5.72 & - & - & N/A & N/A & N/A\\
Aberman & 2.76 & - & - & \underline{2.25} & \textbf{1.15} & \underline{0.08}\\
SAME$^{\dagger}$ & 2.91 & \underline{1.87} & \underline{0.18} & 2.47 & 1.94 & 0.17\\
PALUM$^{\dagger}$ & \underline{2.72} & - & - & 5.67 & - & -\\
Ours$^{\dagger\ddagger}$ & \textbf{1.45} & \textbf{0.72} & \textbf{0.03} & \textbf{1.28} & \underline{0.72} & \textbf{0.05}\\
\bottomrule
\end{tabular}
\label{tab:retargeting_results}
\end{table}

\begin{figure}[t]
  \centering
  \includegraphics[width=0.8\linewidth]{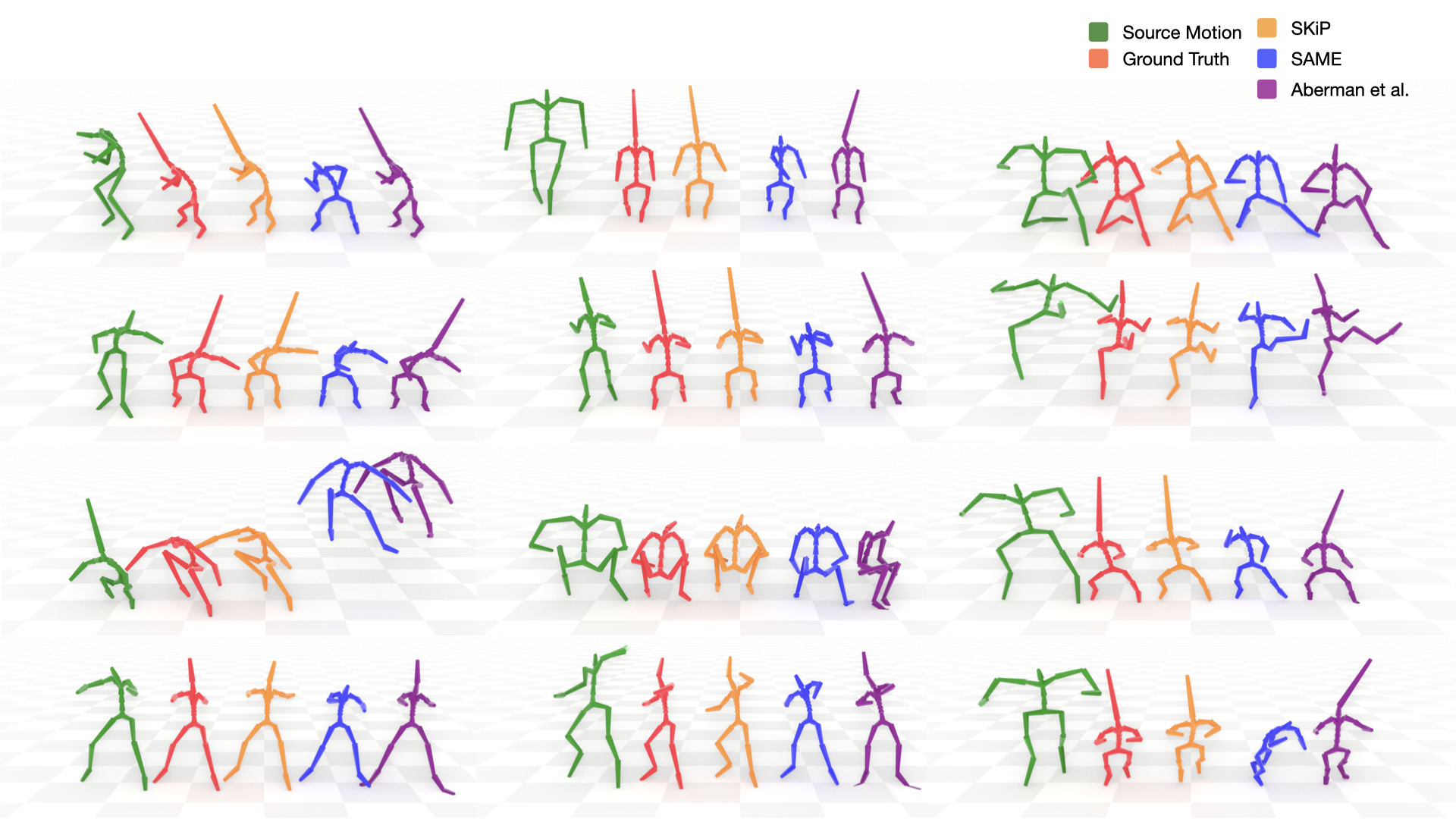}
  \caption{Retargeting across diverse skeletons: The green skeletons are the source motions, red are ground truth retargets from the Mixamo dataset,  the orange are retargets from our method, blue are from SAME and purple from \citet{Aberman2020}.}
  \label{fig:retarget_paper_comparison}
\end{figure}
\subsubsection{Qualitative Measurement}

% \begin{wraptable}{r}{0.5\linewidth}
% \vspace{-0.8em}
% \centering
% \small
% \setlength{\tabcolsep}{3pt}
% \caption{\textbf{Bold} indicates best; * indicates statistical significance against our method (Mann-Whitney U test $p<0.05$).}
% \label{tab:user_study}
% \begin{tabular}{lcc}
% \toprule
% Method & Alignment $\uparrow$ & Quality $\uparrow$ \\
% \midrule
% Aberman & $4.00 \pm 0.05$ & $3.49 \pm 0.06$* \\
% SAME & $3.42 \pm 0.06$* & $3.69 \pm 0.06$* \\
% Ours & $\mathbf{4.06 \pm 0.05}$ & $\mathbf{3.86 \pm 0.05}$ \\
% \midrule
% Ground truth & $4.42 \pm 0.04$* & $4.39 \pm 0.04$* \\
% \bottomrule
% \end{tabular}
% \vspace{-0.5em}
% \end{wraptable}

To corroborate our quantitative results, we conducted a user study in which 37 participants, including 8 professional animators, evaluated retargeted animations from our method, SAME, \citet{Aberman2020}, 
\begin{wraptable}{r}{0.42\textwidth}
  \centering
  \vspace{-0.8\baselineskip}
  \caption{User study. \textbf{Bold}: best; * significant vs.\ ours ($p < 0.05$).}
  \small
  \begin{tabular}{lcc}
    \toprule
    \textbf{Method} & \textbf{Align.} $\uparrow$ & \textbf{Qual.} $\uparrow$ \\
    \midrule
    Aberman          & $4.00 \pm 0.05$            & $3.49 \pm 0.06^*$ \\
    SAME             & $3.42 \pm 0.06^*$           & $3.69 \pm 0.06^*$ \\
    Ours             & $\mathbf{4.06 \pm 0.05}$   & $\mathbf{3.86 \pm 0.05}$ \\
    \midrule
    Ground truth     & $4.42 \pm 0.04^*$           & $4.39 \pm 0.04^*$ \\
    \bottomrule
  \end{tabular}
  \label{tab:user_study}
\end{wraptable} 
and ground truth. Participants viewed each animation individually in random order and rated it on two criteria, alignment to the source motion and perceptual quality, using a 1 (poor) to 5 (excellent) Likert scale. Ground truth serves as an upper bound. While our method closes the gap to ground truth, a perceptual difference remains. A two-sided Mann-Whitney U test ($p < 0.05$) indicates that our 
method significantly outperforms SAME and \citet{Aberman2020} in quality, and SAME in alignment (Table~\ref{tab:user_study}). Qualitative comparisons in Figure~\ref{fig:retarget_paper_comparison}, and~\ref{fig:retarget_paper_extra} in Appendix~\ref{appendix:retargeting} further show coherent retargeting across diverse skeleton topologies.

\subsection{Ablation Study}

We ablate the key architectural and training design choices of our method in Table~\ref{tab:retargeting_results_ablation}. Replacing the Transformer with a GAT~\citep{velickovic2018gat} substantially degrades reconstruction and retargeting performance, resulting in very large GJP errors ($891.96/927.38$ for Intra/Cross). The extreme magnitude of these errors is primarily driven by inaccurate global root-trajectory prediction ($>233$ cm), although local motion reconstruction also deteriorates. The training curves reveal that the GAT overfits to the training skeletons, failing to generalize to the joint modeling of local pose and global trajectory required for retargeting. Removing positional encoding (PE) similarly causes the model to fail, while replacing our multiplicative PE with an additive formulation increases GJP by $42\%$ on Intra ($1.45 \rightarrow 2.06$) and $57\%$ on Cross ($1.28 \rightarrow 2.01$). Finally, removing our augmentations substantially degrades generalization, increasing GJP to $3.92/2.66$. Together, these results demonstrate the importance of the Transformer architecture, multiplicative structural encoding, and augmentation strategy. Ablations of the individual training losses are provided in Table~\ref{tab:retargeting_loss_ablation} in Appendix~\ref{appendix:loss_ablation}.

\begin{table}[h]
\caption{Ablation Study on the Mixamo dataset. We evaluate the impact of removing key components on Intra- and Cross-retargeting performance.
\textbf{Bold} indicates best; \underline{underline} indicates second best.}
\centering
\begin{tabular}{lcccccc}
\toprule
& \multicolumn{3}{c}{Intra} & \multicolumn{3}{c}{Cross} \\
\cmidrule(lr){2-4} \cmidrule(lr){5-7}
Method & GJP & Jerk & PART & GJP & Jerk & PART \\
\midrule
Ours Full& \textbf{1.45} & \textbf{0.72} & \textbf{0.03} & \textbf{1.28} & \textbf{0.72} & \textbf{0.05}\\
with GAT & 891.96 & 30.07 & 0.38 & 927.38 & 31.43 & 0.36\\
w/o PE & 62.98 & 0.01 & 0.57 & 62.98 & 0.01 & 0.60\\
additive PE & \underline{2.06} & \underline{0.70} & 0.05 & \underline{2.01} & \textbf{0.72} & \underline{0.06}\\
w/o aug. & 3.92 & 0.63 & \underline{0.04} & 2.66 & \underline{0.64} & \textbf{0.05}\\
\bottomrule
\end{tabular}
\label{tab:retargeting_results_ablation}
\end{table}

\section{Summary, Limitations, and Future Work}
In this work, we introduce a unified, unsupervised Transformer Autoencoder for motion retargeting across arbitrary skeletal topologies. While prior transformer-based attempts such as PALUM fail when retargeting between characters with differing skeletal structures, our learnable skeletal graph flattening with multiplicative positional encodings is what makes transformers effective for this task. Unlike~\citet{Aberman2020} and SAME, our single model requires neither topology-specific components nor paired training data, and avoids delta-based error accumulation by predicting the global root trajectory directly. Evaluated entirely in a zero-shot setting, our method outperforms all existing state-of-the-art methods both quantitatively and qualitatively.

% \subsection{Limitations and Future Work}

Despite these advancements, several limitations remain. Our model assumes a T-pose rest configuration, requiring preprocessing of input motion data. Additionally, our loss functions focus on skeletal kinematics and do not account for surface geometry, which can lead to self-penetration for characters with voluminous meshes or misalignment between the skeleton and the mesh, such as shoulders dropping when the mesh is applied. Our model is also designed for human-like skeletal structures and struggles with characters that have a large number of joints from non-standard body parts, such as capes or other protruding elements. Future work could address these limitations by introducing mesh-aware constraints and extending the model to handle more diverse skeletal configurations, and explore leveraging the learned latent space for large-scale motion modelling across diverse datasets.

\subsection*{AI use statement}

In this work, we used generative AI tools for none of the tasks with required disclosure. We have \textbf{not used} generative AI tools for help to develop theoretical models or conceptual frameworks, formulate mathematical claims, provide critical ingredients for proving mathematical claims, assist in the writing of proofs, propose or refine hypotheses, design or provide feedback on research  methodology or experiments, implement methods, clean and reformat dataset, support qualitative and thematic data analysis, or interpret results,
and generating synthetic data sets, or assist with translation are not applicable to this work. We \textbf{used} generative AI tools for editing the research paper to improve readability and formatting references. We have reviewed all AI-assisted work. All AI-assisted edits were manually reviewed by the authors for correctness and consistency, and all generated BibTeX entries were verified against the original sources. We take responsibility for the final content of this work, including text, claims or artifacts produced with the aid of generative AI.

%(This section is \textbf{required} and does not count toward the page limit.)

%In this work, we used generative AI tools for [tasks with required disclosure].
%[Elaborate. For example, “we
%checked LLM-generated research ideas for potential plagiarism through a manual
%literature survey”, “LLM-generated code was verified and tested for correctness
%by 2 authors”, etc.]. We take responsibility for the final content of this work,
%including text, claims or artifacts produced with the aid of generative AI.

%See the ICLR 2027 AI Policy for Authors for more details. This statement should
%not be more than 1 page.

\subsection*{Reproducibility statement}

To ensure the reproducibility of our work, we have provided detailed information on our method and experimental setups. We will discuss the respective details below.

\textbf{Our method and architecture} In addition to the details presented in the main text (Section~\ref{sec:methods})(Sec. 3), we provided detailed descriptions of the model architecture in (Section~\ref{app:model_architecture}), the hyperparameter settings (Section~\ref{app:hyperparameter}), and the loss functions and their weightings. Furthermore, we provided detailed descriptions of all data augmentation strategies used in our work (Section~\ref{app:data_augment}), including skeleton augmentation (Section~\ref{app:skeleton_augmentation}), bone scaling (Section~\ref{app:bone_Scaling_aug}), global translation (Section~\ref{app:global_translation_augmentation}), and rest pose augmentation (Section~\ref{app:restpose_augment}).

\textbf{Experiments} In addition to the details provided in the main text (Section~\ref{sec:experiments}), we provided extensive additional information (Section~\ref{sec:architecture_details}). Specifically, we detail the data preparation and evaluation protocols (Section~\ref{app_sec:data_prep}), and describe the training protocol and hardware setup (Section~\ref{app:training_protocol}).

\textbf{Code Availability} The complete implementation, along with pretrained models and detailed instructions for reproducing our results, is publicly available at \url{https://github.com/sinzlab/retarget}. The repository includes guidelines for setting up the environment and executing the training pipeline.

%\subsubsection*{Author Contributions}
%If you'd like to, you may include  a section for author contributions as is done
%in many journals. This is optional and at the discretion of the authors.

\subsubsection*{Acknowledgments}
FHS is supported by the European Research Council (ERC) under the European Union's Horizon Europe research and innovation programme (Grant agreement No. 101171526) and by the German Research Foundation (DFG) SFB 1456, Mathematics of Experiment -- Project-ID 432680300.

\clearpage
\bibliography{iclr2027_conference}

@String{Computing = "Computing" }

@String{Computer = "{IEEE} Computer" }

@article{Gleicher1998,
  author = {Michael Gleicher},
  title = {Retargetting Motion to New Characters},
  journal = {Proceedings of SIGGRAPH},
  year = {1998},
  pages = {33–42},
  publisher = {ACM},
}

@article{Tak2005,
  author    = {Tak, Seyoon and Ko, Hyeong-Seok},
  title     = {A Physically-Based Motion Retargeting Filter},
  journal   = {ACM Transactions on Graphics (TOG)},
  volume    = {24},
  number    = {1},
  pages     = {98--117},
  year      = {2005},
  publisher = {ACM},
  doi       = {10.1145/1037957.1037963},
}

@inproceedings{Lee1999,
  author    = {Jehee Lee and Sung Yong Shin},
  title     = {A Hierarchical Approach to Interactive Motion Editing for Human-like Figures},
  booktitle = {Proceedings of the 26th Annual Conference on Computer Graphics and Interactive Techniques (SIGGRAPH '99)},
  pages     = {39--48},
  year      = {1999},
  publisher = {ACM Press/Addison-Wesley Publishing Co.},
}

@article{Choi2000,
  author    = {Kwang-Jin Choi and Hyeong-Seok Ko},
  title     = {Online Motion Retargetting},
  journal   = {Journal of Visualization and Computer Animation},
  volume    = {11},
  number    = {5},
  pages     = {223--235},
  year      = {2000},
  publisher = {Wiley},
}

@inproceedings{Villegas2018,
  title={Neural Kinematic Networks for Unsupervised Motion Retargetting},
  author={Ruben Villegas and Jimei Yang and Duygu Ceylan and Honglak Lee},
  booktitle={Proceedings of the IEEE Conference on Computer Vision and Pattern Recognition (CVPR)},
  year={2018},
  pages={8639--8648},
}

@article{Aberman2020,
  title     = {Skeleton-Aware Networks for Deep Motion Retargeting},
  author    = {Aberman, Kfir and Li, Peizhuo and Lischinski, Dani and Sorkine-Hornung, Olga and Cohen-Or, Daniel and Chen, Baoquan},
  journal   = {ACM Transactions on Graphics (TOG)},
  volume    = {39},
  number    = {4},
  pages     = {120},
  year      = {2020},
}

@inproceedings{Lee2023SAME,
  title     = {SAME: Skeleton-Agnostic Motion Embedding for Character Animation},
  author    = {Lee, Sunmin and Kang, Taeho and Park, Jungnam and Lee, Jehee and Won, Jungdam},
  booktitle = {ACM SIGGRAPH ASIA 2023 Conference Proceedings},
  year      = {2023},
  url       = {https://sunny-codes.github.io/assets/pdf/SAME.pdf}
}

@misc{autodesk_motionbuilder,
  title = {MotionBuilder - A 3D Character Animation Software},
  author = {Autodesk},
  year = {2021},
  url = {https://www.autodesk.com/products/motionbuilder/overview}
}

@inproceedings{holden2015learning,
  title     = {Learning motion manifolds with convolutional autoencoders},
  author    = {Holden, Daniel and Saito, Jun and Komura, Taku and Joyce, Thomas},
  booktitle = {SIGGRAPH Asia 2015 Technical Briefs},
  pages     = {1--4},
  year      = {2015},
  organization = {ACM},
}

@article{aberman2019learning,
  title={Learning Character-Agnostic Motion for Motion Retargeting in 2D},
  author={Aberman, Kfir and Wu, Rundi and Lischinski, Dani and Chen, Baoquan and Cohen-Or, Daniel},
  journal={ACM Transactions on Graphics (TOG)},
  volume={38},
  number={4},
  pages={75},
  year={2019},
  publisher={ACM},
}

@inproceedings{athanasiou2022teach,
  title     = {TEACH: Temporal Action Composition for 3D Humans},
  author    = {Athanasiou, Nikos and Petrovich, Mathis and Black, Michael J and Varol, G{\"u}l},
  booktitle = {Proceedings of the International Conference on 3D Vision (3DV)},
  year      = {2022},
  organization = {IEEE},
  url       = {https://teach.is.tue.mpg.de}
}

@article{starke2022deepphase,
  title     = {DeepPhase: Periodic Autoencoders for Learning Motion Phase Manifolds},
  author    = {Starke, Sebastian and Mason, Ian and Komura, Taku},
  journal   = {ACM Transactions on Graphics (TOG)},
  volume    = {41},
  number    = {4},
  pages     = {136:1--136:13},
  year      = {2022},
  publisher = {ACM},
}

@inproceedings{mahmood2019amass,
  title     = {AMASS: Archive of Motion Capture as Surface Shapes},
  author    = {Mahmood, Naureen and Ghorbani, Nima and Troje, Nikolaus F. and Pons-Moll, Gerard and Black, Michael J.},
  booktitle = {Proceedings of the IEEE/CVF International Conference on Computer Vision (ICCV)},
  year      = {2019},
  url       = {https://amass.is.tue.mpg.de}
}

@inproceedings{loper2015smpl,
  title     = {SMPL: A Skinned Multi-Person Linear Model},
  author    = {Loper, Matthew and Mahmood, Naureen and Romero, Javier and Pons-Moll, Gerard and Black, Michael J.},
  booktitle = {ACM Transactions on Graphics (TOG)},
  volume    = {34},
  number    = {6},
  pages     = {248},
  year      = {2015},
  publisher = {ACM},
  url       = {http://smpl.is.tue.mpg.de}
}

@inproceedings{pavlakos2019expressive,
  title     = {Expressive Body Capture: 3D Hands, Face, and Body from a Single Image},
  author    = {Pavlakos, Georgios and Choutas, Vasileios and Ghorbani, Nima and Bolkart, Timo and Osman, Ahmed A. A. and Tzionas, Dimitrios and Black, Michael J.},
  booktitle = {Proceedings of the IEEE/CVF Conference on Computer Vision and Pattern Recognition (CVPR)},
  pages     = {10975--10985},
  year      = {2019},
}

@inproceedings{rempe2021humor,
  title     = {HuMoR: 3D Human Motion Model for Robust Pose Estimation},
  author    = {Rempe, Davis and Birdal, Tolga and Hertzmann, Aaron and Yang, Jimei and Sridhar, Srinath and Guibas, Leonidas J.},
  booktitle = {Proceedings of the IEEE/CVF International Conference on Computer Vision (ICCV)},
  pages     = {11488--11499},
  year      = {2021},
}

@article{chen2022learning,
  title     = {Learning Variational Motion Prior for Video-based Motion Capture},
  author    = {Chen, Xin and Su, Zhuo and Yang, Lingbo and Cheng, Pei and Xu, Lan and Fu, Bin and Yu, Gang},
  journal   = {arXiv preprint arXiv:2210.15134},
  year      = {2022},
  url       = {https://arxiv.org/abs/2210.15134}
}

@inproceedings{raab2023single,
  title     = {Single Motion Diffusion},
  author    = {Raab, Sigal and Leibovitch, Inbal and Tevet, Guy and Arar, Moab and Bermano, Amit H. and Cohen-Or, Daniel},
  booktitle = {International Conference on Learning Representations (ICLR)},
  year      = {2024},
  url       = {https://arxiv.org/abs/2302.05905},
}

@inproceedings{yuan2022physdiff,
  title     = {PhysDiff: Physics-Guided Human Motion Diffusion Model},
  author    = {Yuan, Ye and Song, Jiaming and Iqbal, Umar and Vahdat, Arash and Kautz, Jan},
  booktitle = {Proceedings of the IEEE/CVF International Conference on Computer Vision (ICCV)},
  pages     = {16010--16021},
  year      = {2023},
  doi       = {10.1109/ICCV51070.2023.01467},
}

@inproceedings{hamilton2017inductive,
  title     = {Inductive Representation Learning on Large Graphs},
  author    = {Hamilton, William L. and Ying, Rex and Leskovec, Jure},
  booktitle = {Proceedings of the 31st International Conference on Neural Information Processing Systems (NeurIPS)},
  pages     = {1024--1034},
  year      = {2017},
  publisher = {Curran Associates, Inc.},
  url       = {https://arxiv.org/abs/1706.02216}
}

@inproceedings{zhu2017unpaired,
  title     = {Unpaired Image-to-Image Translation using Cycle-Consistent Adversarial Networks},
  author    = {Zhu, Jun-Yan and Park, Taesung and Isola, Phillip and Efros, Alexei A.},
  booktitle = {Proceedings of the IEEE International Conference on Computer Vision (ICCV)},
  pages     = {2223--2232},
  year      = {2017},
  url       = {https://arxiv.org/abs/1703.10593}
}

@inproceedings{goodfellow2014generative,
  title     = {Generative Adversarial Nets},
  author    = {Goodfellow, Ian J. and Pouget-Abadie, Jean and Mirza, Mehdi and Xu, Bing and Warde-Farley, David and Ozair, Sherjil and Courville, Aaron and Bengio, Yoshua},
  booktitle = {Advances in Neural Information Processing Systems (NeurIPS)},
  pages     = {2672--2680},
  year      = {2014},
  url       = {https://arxiv.org/abs/1406.2661}
}

@article{harvey2020robust,
  author    = {Félix G. Harvey and Mike Yurick and Derek Nowrouzezahrai and Christopher Pal},
  title     = {Robust Motion In-Betweening},
  journal   = {ACM Transactions on Graphics (Proceedings of ACM SIGGRAPH)},
  publisher = {ACM},
  volume    = {39},
  number    = {4},
  year      = {2020}
}

@misc{accad,
  title  = {ACCAD Motion Capture Database},
  howpublished = {\url{https://accad.osu.edu/research/mocap/mocap_data.html}},
  note   = {Accessed: 2025-05-19}
}

@inproceedings{trumble2017totalcapture,
  title     = {Total Capture: 3D Human Pose Estimation Fusing Video and Inertial Sensors},
  author    = {Trumble, Matthew and Gilbert, Andrew and Malleson, Charles and Hilton, Adrian and Collomosse, John},
  booktitle = {Proceedings of the British Machine Vision Conference (BMVC)},
  year      = {2017},
  publisher = {BMVA Press},
}

@article{holden2017phase,
  title   = {Phase-Functioned Neural Networks for Character Control},
  author  = {Holden, Daniel and Komura, Taku and Saito, Jun},
  journal = {ACM Transactions on Graphics (TOG)},
  volume  = {36},
  number  = {4},
  pages   = {42:1--42:13},
  year    = {2017},
}

@misc{sfu2011dataset,
  title        = {SFU Motion Capture Dataset},
  author       = {{Simon Fraser University}},
  year         = {2011},
  howpublished = {\url{http://mocap.cs.sfu.ca/}},
  note         = {Accessed: 2025-05-19}
}

@misc{kobayashi2023motion,
  title={Motion Capture Dataset for Practical Use of AI-based Motion Editing and Stylization},
  author={Makito Kobayashi and Chen-Chieh Liao and Keito Inoue and Sentaro Yojima and Masafumi Takahashi},
  year={2023},
  eprint={2306.08861},
  archivePrefix={arXiv},
  primaryClass={cs.CV}
}

@misc{adobe2021mixamo,
  author = {Adobe},
  title = {Mixamo Animation Dataset},
  year = {2021},
  howpublished = {\url{https://www.mixamo.com}}
}

@inproceedings{gat2025anytop,
  title     = {AnyTop: Character Animation Diffusion with Any Topology},
  author    = {Gat, Inbar and Raab, Sigal and Tevet, Guy and Reshef, Yuval and Bermano, Amit H. and Cohen-Or, Daniel},
  booktitle = {Proceedings of the ACM SIGGRAPH Conference},
  year      = {2025},
  url       = {https://anytop2025.github.io/Anytop-page/}
}

@inproceedings{zhou2019continuity,
  title     = {On the Continuity of Rotation Representations in Neural Networks},
  author    = {Zhou, Yi and Barnes, Connelly and Lu, Jingwan and Yang, Jimei and Li, Hao},
  booktitle = {Proceedings of the IEEE/CVF Conference on Computer Vision and Pattern Recognition (CVPR)},
  pages     = {5738--5746},
  year      = {2019},
}

@inproceedings{bruderlin1995motion,
  author    = {Armin Bruderlin and Lance Williams},
  title     = {Motion Signal Processing},
  booktitle = {Proceedings of SIGGRAPH 1995},
  year      = {1995},
  pages     = {97--104},
  publisher = {ACM},
}

@inproceedings{holden2016deep,
  author    = {Daniel Holden and Jun Saito and Taku Komura},
  title     = {A Deep Learning Framework for Character Motion Synthesis and Editing},
  booktitle = {ACM Transactions on Graphics (SIGGRAPH 2016)},
  year      = {2016},
  volume    = {35},
  number    = {4},
  pages     = {138:1--138:11},
}

@inproceedings{pavllo20193dhpe,
  author    = {Dario Pavllo and Christoph Feichtenhofer and David Grangier and Michael Auli},
  title     = {3D Human Pose Estimation in Video with Temporal Convolutions and Semi-Supervised Training},
  booktitle = {Proceedings of the IEEE/CVF Conference on Computer Vision and Pattern Recognition (CVPR)},
  year      = {2019},
  pages     = {7753--7762},
}

@inproceedings{petrovich2021action,
  author    = {Maxim Petrovich and Michael J. Black and Christoph Lassner},
  title     = {Action-Conditioned 3D Human Motion Synthesis with Transformer VAE},
  booktitle = {Proceedings of the IEEE/CVF International Conference on Computer Vision (ICCV)},
  year      = {2021},
  pages     = {10981--10991},
}

@inproceedings{tevet2022human,
  title     = {Human Motion Diffusion Model},
  author    = {Tevet, Guy and Raab, Sigal and Gordon, Brian and Shafir, Yonatan and Cohen-Or, Daniel and Bermano, Amit H.},
  booktitle = {International Conference on Learning Representations (ICLR)},
  year      = {2023},
  url       = {https://arxiv.org/abs/2209.14916},
}

@article{liu2026palum,
  title     = {PALUM: Part-based Attention Learning for Unified Motion Retargeting},
  author    = {Liu, Siqi and Wang, Maoyu and Dai, Bo and Lu, Cewu},
  journal   = {arXiv preprint arXiv:2601.07272},
  year      = {2026}
}

@article{mason2022local,
author = {Mason, Ian and Starke, Sebastian and Komura, Taku},
title = {Real-Time Style Modelling of Human Locomotion via Feature-Wise Transformations and Local Motion Phases},
year = {2022},
publisher = {Association for Computing Machinery},
address = {New York, NY, USA},
volume = {5},
number = {1},
journal = {Proceedings of the ACM on Computer Graphics and Interactive Techniques},
month = {may},
articleno = {6}
}

@article{chen2025motion2motion,
  title={Motion2Motion: Cross-topology Motion Transfer with Sparse Correspondence},
  author={Chen, Ling-Hao and Zhang, Yuhong and Yin, Zixin and Dou, Zhiyang and Chen, Xin and Wang, Jingbo and Komura, Taku and Zhang, Lei},
  journal={ACM SIGGRAPH Asia 2025 Conference Proceedings},
  year={2025}
}

@article{jang2022motionpuzzle,
  title={Motion Puzzle: Arbitrary Motion Style Transfer by Body Part},
  author={Jang, Deok-Kyeong and Park, Soomin and Lee, Sung-Hee},
  journal={ACM Transactions on Graphics (TOG)},
  volume={41},
  number={3},
  pages={1--16},
  year={2022},
  publisher={ACM}
}

@misc{cmu2003mocap,
  title = {CMU Graphics Lab Motion Capture Database},
  author = {{Carnegie Mellon University}},
  year = {2003},
  howpublished = {\url{http://mocap.cs.cmu.edu/}},
  note = {Accessed: 2025-05-19}
}

@inproceedings{velickovic2018gat,
  title={Graph Attention Networks},
  author={Veli{\v{c}}kovi{\'{c}}, Petar and Cucurull, Guillem and Casanova, Arantxa and Romero, Adriana and Li{\`{o}}, Pietro and Bengio, Yoshua},
  booktitle={International Conference on Learning Representations (ICLR)},
  year={2018}
}
\bibliographystyle{iclr2027_conference}

\clearpage

\appendix
\section{\MakeUppercase{Architecture Details}}
\label{sec:architecture_details}

This section details our transformer autoencoder architecture, including model specifications, hyperparameter settings, loss function weighting, training protocol, and code availability.

\subsection{Model Architecture}\label{app:model_architecture}

Our autoencoder is composed of an encoder and a decoder. The encoder consists of a linear projection layer followed by a transformer encoder. The positional encoding in the encoder is learned using a GraphSAGE layer. Similarly, the decoder is built with a transformer encoder and a linear decoding layer, and it employs a GraphSAGE layer to learn its positional encoding. Furthermore it introduces two learnable parameters, used for the latent representation of the pose and root trajectory separately. Also for the discriminator network, which uses only the transformer encoder part, we introduce a learnable parameter, used for the classification. Table~\ref{tab:architecture_specs} summarizes these components.

\begin{table}[ht]
\centering
\caption{Model Architecture Specifications}
\label{tab:architecture_specs}
\begin{tabular}{ll}
\toprule
\textbf{Component} & \textbf{Specification} \\
\midrule
\textbf{Encoder} &  \\
Linear Projection & Embedding dimension: 128 \\
Transformer Encoder & Layers: 4, Attention Heads: 8\\ & FeedForward: 512\\
Positional Encoding & GraphSAGE 2 layers \\
\midrule
\textbf{Decoder} &  \\
Transformer Encoder & Layers: 4, Attention Heads: 8 \\ & FeedForward: 512\\
Linear Decoding Dimensions & 6 (For joints), 3 (For Trajectory) \\
Positional Encoding & GraphSAGE 2 layers \\
\midrule
\textbf{Adversarial Discriminator} &  \\
Linear Projection & Embedding dimension: 128 \\
Transformer Encoder & Layers: 2, Attention Heads: 2, \\ & FeedForward: 64\\
Positional Encoding & GraphSAGE 2 layers \\
Linear Decoding Dimensions & 1 \\
\midrule
\bottomrule
\end{tabular}
\end{table}

\subsection{Hyperparameter Settings}\label{app:hyperparameter}

The training process employs the hyperparameters listed in Table~\ref{tab:hyperparams}. These settings were chosen based on extensive experimentation to ensure stable convergence and robust performance. It is to note, that for the adversarial discriminator optimizer, no scheduler was used.

\begin{table}[ht]
\centering
\caption{Hyperparameter Settings}
\label{tab:hyperparams}
\begin{tabular}{ll}
\toprule
\textbf{Hyperparameter} & \textbf{Value} \\
\midrule
Batch Size & $128 \times 8$ \\
Weight Decay & $1 \times 10^{-4}$\\
Learning Rate & $1 \times 10^{-3}$ \\
Learning Rate Scheduler & Cosine Annealing with Restarts\\
Optimizer & AdamW \\
\bottomrule
\end{tabular}
\end{table}

\subsection{Loss Function and Weighting}\label{app_sec:loss_func_weighting}

\subsubsection{Reconstruction Loss}
The primary goal is to accurately reconstruct a globally scaled input motion, while for the cycled back skeleton we reconstruct the unscaled input motion.
This is achieved by minimizing a reconstruction loss that is composed of multiple components.
First, the \textit{position loss} minimizes the Euclidean distance between the ground truth positions $p$ and the reconstructed positions $\hat{p}$ (\ref{eq:position_loss}).
Additionally, the \textit{root children position loss} minimizes the Euclidean distance between the ground truth positions $p$ and the reconstructed positions $\hat{p}$ of the children joints of the root joint(\ref{eq:root_children_position_loss}).
Next, the \textit{rotation loss} employs a geodesic distance metric to quantify the angular discrepancy between the ground truth rotations $q$ and the predicted rotations $\hat{q}$ (\ref{eq:rotation_loss}).
The \textit{trajectory loss} minimizes the Euclidean distance between the ground truth trajectory $r^t$ and its reconstruction $\hat{r}^t$(\ref{eq:traj_loss}).
\begin{align}
    \mathcal{L}_\text{rec}(\mathcal{X}^t, \hat{\mathcal{X}}^t)
    &= \lambda_\text{pos} \frac{1}{J} \sum_{j=1}^{J} \| p_j^t - \hat{p}_j^t \| \label{eq:position_loss} \\
    &+ \lambda_\text{child} \frac{1}{J} \sum_{j=1}^{J_\text{child}} \| p_j^t - \hat{p}_j^t \| \label{eq:root_children_position_loss} \\
    &+ \lambda_\text{rot}\frac{1}{J} \sum_{j=1}^{J} d_g(q_j^t, \hat{q}_j^t) \label{eq:rotation_loss} \\
    &+ \lambda_\text{traj}\| r^t - \hat{r}^t \|\label{eq:traj_loss}
\end{align}

\subsubsection{Temporal Smoothing}
Since our model reconstructs frame by frame individually, we regularize the temporal derivatives of the motion by introducing the \textit{velocity loss}, which minimizes the Euclidean distance between the reconstructed velocity $\hat{v}_t$ and the ground truth velocity $v_t$ (\ref{eq:velocity_loss}).
Moreover, we also penalize discrepancies in joint jerks by adding a \textit{jerk loss}, which minimizes the Euclidean distance between reconstructed jerk $\hat{a}^t - \hat{a}^{t-1}$ and ground truth jerk $a^t - a^{t-1}$ (\ref{eq:jerk_loss}). Here, $a$ describes the acceleration and $t_\text{frame}$ is the frame time.
\begin{align}
    \mathcal{L}_\text{temp}(\mathcal{X}^t, \hat{\mathcal{X}}^t)
        &= \lambda_\text{vel}\sum_{j=1}^{J} \Big\| v_j^t -  \hat{v}_j^t \Big\| \label{eq:velocity_loss} \\
        &+ \frac{\lambda_\text{jerk}}{t_\text{frame}^3} \sum_{j=1}^{J}\Big\| (a_j^{t} - a_j^{t-1}) - (\hat{a}^t - \hat{a}^{t-1}) \Big\| \label{eq:jerk_loss}
\end{align}

\subsubsection{Contact Loss}
One common issue with deep learning based retargeting methods is foot sliding and ground penetration. In order to tackle the foot sliding, we introduce a \textit{contact position loss}. This minimizes the distance of the reconstructed height $\hat{y}$ to the ground. It is only applied on the joints, for which the ground truth joints are in contact with the ground (\ref{eq:contact_pos}). To generate these ground truth contact labels for training, a joint is considered to be in contact if its height falls below a threshold referred to as the \textit{contact position}. The contact position is defined per animation by computing, for each frame, the minimum joint height and taking the 5th percentile of these values as the final threshold.
We emphasize that this heuristic is strictly a data pre-processing step used to derive supervision signals; the contact labels are not required during inference.

Furthermore, we introduce a \textit{contact velocity loss}, which forces the reconstructed velocity $\hat{v}$ of the ground contact joints to be zero, so that the feet are not sliding along the ground (\ref{eq:contact_vel}). 
To penalize ground penetration, we introduce a \textit{ground penetration loss}, that penalizes joints penetrating the ground plane by enforcing upward correction toward zero penetration (\ref{eq:ground_pen}).

\begin{align}
    \mathcal{L}_\text{cntct}(\mathcal{X}^t, \hat{\mathcal{X}}^t)
        &= \lambda_\text{cntct\_pos}\sum_{j=1}^{J_\text{cntct}} \Big\|\hat{y}_j^t \Big\| \label{eq:contact_pos} \\
        &+ \lambda_\text{cntct\_vel} \sum_{j=1}^{J_\text{cntct}}\Big\|\hat{v}_j^t \Big\| \label{eq:contact_vel} \\
        &+ \lambda_\text{grnd\_pen} \sum_{j=1}^{J}\Big\|\mathrm{min}(\hat{y}_j^t ,0)\Big\| \label{eq:ground_pen}
\end{align}

\subsubsection{End Effector Loss}
Different sized skeletons should have the same normalized velocities while performing the same motion. In order to enforce that condition, we introduce an \textit{end-effector loss} like \citep{Aberman2020}. This forces the velocities of the end effectors $v$ normalized by its path length $h$ to be the same between skeletons A and B. The skeletons A and B have to share the same number of end-effectors.
\begin{align*}
    \mathcal{L}_\text{ee}(\mathcal{X}_A^t, \hat{\mathcal{X}}_B^t)
        &= \lambda_\text{ee}\sum_{j=1}^{J_\text{ee}} \left\|\frac{v_{A,j}^t}{h_A} -  \frac{\hat{v}_{B,j}^t}{h_B} \right\|
\end{align*}

\subsubsection{Latent Consistency}
Stability in the latent representation is enforced by minimizing the Euclidean distance between the latent encoding $z_p$ and its reconstructed version $\hat{z}_p$.
\[
\mathcal{L}_\text{latent}(\mathcal{X}, \hat{\mathcal{X}}) =\lambda_\text{zpose} \| z_p - \hat{z}_p \|
\]

\subsubsection{Adversarial Loss}
Since we have no access to the ground truth retargeting motion between two skeletons, we introduce an adversarial loss. This ensures more realistic looking motions, especially for the retargeted motions. In order to train with an adversarial loss, we introduce a separate discriminator network $\mathcal{D}$. The adversarial loss consists of two parts. The first part is the discriminator part, which classifies if the input is real (Equation \ref{eq:discriminator_real}) or fake (Equation \ref{eq:discriminator_fake}). For that, only the weights of the discriminator network is updated.

The second part is the generator part, which tries to make the discriminator predict the auto-encoder output as real sample (Equation \ref{eq:generator}). Here, only the auto-encoder weights are updated. 
\begin{align}
    \mathcal{L}_\text{adv}(\mathcal{X}^t, \hat{\mathcal{X}}^t)
        &= \lambda_\text{disc} E_{i \sim \mathcal{M}_A}\left[\Big\| 1 - \mathcal{D}(\mathcal{X}^t_i,\mathcal{S}_i) \Big\|\right]\label{eq:discriminator_real} \\
        &+\lambda_\text{disc} E_{i \sim \mathcal{M}_B}\left[\Big\| \mathcal{D}(\hat{\mathcal{X}}^t_i,\mathcal{S}_i) \Big\|\right]\label{eq:discriminator_fake} \\
        &+\lambda_\text{gen} E_{i \sim \mathcal{M}_B}\left[\Big\|1- \mathcal{D}(\hat{\mathcal{X}}^t_i,\mathcal{S}_i) \Big\|\right]\label{eq:generator} 
\end{align}

\subsubsection{Loss Function Weighting}

Our training objective balances multiple loss terms. The weighting coefficients for these losses are provided in Table~\ref{tab:loss_weights}. For example, the positional loss and the rotation loss are weighted by \(\lambda_{\text{pos}}\) and \(\lambda_{\text{rot}}\), respectively. Notably, the corresponding weights for cycle consistency and augmentation consistency is the same. One example is, that \(\lambda_{\text{pos}}\) is the same in $\mathcal{L}_{\mathrm{cyc. cons.}}$ and $\mathcal{L}_{\mathrm{aug. cons.}}$. 

\clearpage

\begin{table}[ht]
\centering
\caption{Loss Function Weighting Coefficients}
\label{tab:loss_weights}
\begin{tabular}{ll}
\toprule
\textbf{Loss} & \textbf{Weight} \\
\midrule
\(\lambda_{\text{pos}}\) & $10^{2}$ \\
\(\lambda_{\text{child}}\) & $10^{2}$ \\
\(\lambda_{\text{rot}}\) & 5 \\
\(\lambda_{\text{jerk}}\) & $10^{-5}$ \\
\(\lambda_{\text{traj}}\) & 10 \\
\(\lambda_{\text{vel}}\) & 1 \\
\(\lambda_{\text{trans}}\) & 1 \\
\(\lambda_{\text{cntct\_pos}}\) & 1 \\
\(\lambda_{\text{cntct\_vel}}\) & 1 \\
\(\lambda_{\text{grnd\_pen}}\) & 0.01 \\
\(\lambda_{\text{ee}}\) & 10 \\
\(\lambda_{\text{zpose}}\) & 0.1 \\
\(\lambda_{\text{disc}}\) & 0.1 \\
\(\lambda_{\text{gen}}\) & 1 \\
% Additional loss components can be added here
\bottomrule
\end{tabular}
\end{table}

\subsection{Technical Details}

\paragraph{Inference Latency.} We measure per-frame latency for a $26 \rightarrow 26$ joint configuration. On CPU, our model runs at $2.46 \pm 0.02$ ms per frame, and on an NVIDIA A100 GPU at $0.12 \pm 0.01$ ms per frame.

\paragraph{Computational Cost.} For a single frame with $26 \rightarrow 26$ joints, our model requires $1.82$ MFLOPs.

\subsection{Data Preparation}
\label{app_sec:data_prep}

To train our autoencoder for per-frame motion retargeting across diverse skeletons, we leverage skeleton-invariant pose embeddings. For this purpose, we use publicly available motion capture datasets, including LaFAN1~\citep{harvey2020robust}, ACCAD~\citep{accad}, TotalCapture~\citep{trumble2017totalcapture}, PFNN~\citep{holden2017phase}, the SFU dataset~\citep{sfu2011dataset}, and CMU~\citep{cmu2003mocap}. These datasets provide a rich variety of motions and skeletal configurations. SAME~\citeyearpar{Lee2023SAME} uses the same base datasets, except for CMU, but additionally includes approximately 13 hours of motion across $\sim$160 different skeletons synthesized via MotionBuilder~\citep{autodesk_motionbuilder} to generate paired ground-truth retargeting data. Aberman et al.~\citeyearpar{Aberman2020} and PALUM~\citeyearpar{liu2026palum} train directly on Mixamo, with the evaluation skeletons held out but the motions already seen during training. Importantly, our method trains entirely unsupervised, requiring neither paired retargeting data nor any Mixamo motion sequences, making our zero-shot evaluation more challenging than prior work.

Before passing the data into the autoencoder for training, we propose a data preprocessing step, which includes scaling of the skeleton $\mathcal{S}$. Assume $h$ is the largest distance between two joints in the rest pose, which for human skeletons is usually the height. The scaling factor of the skeleton $\mathcal{S}$ is then defined as $1/h$. For consistency reasons, also the (previous-) joint positions and root trajectories are scaled by the same factor. In addition to the previous preprocessing, we introduce a feet-grounding step to fix floating-foot artifacts in animation datasets. For each animation, we compute the contact position, which defines when a joint is in contact with the ground. This is calculated by taking the minimum joint height per frame and using the 5th percentile as a robust, scale-invariant threshold. Joint heights in each frame are then translated by the negative contact position, ensuring the animation is properly grounded. This preprocessing is applied on a per-animation basis.

Furthermore, augmentation strategies such as random scaling of the joint lengths, global translations, and joint removal are employed to improve generalization and enables to achieve the demanded invariance. Unlike the preprocessing steps applied before training, these augmentations are applied dynamically during every training step

For testing and evaluation, we use the Bandai-Namco~\citep{kobayashi2023motion} dataset, the 100Style~\citep{mason2022local} dataset and a subset of the Mixamo~\citep{adobe2021mixamo} dataset. Only the Mixamo dataset is used for the Cross and Intra retargeting evaluations, following the evaluation protocols of previous motion retargeting methods, including Aberman et al.~\citeyearpar{Aberman2020} and SAME~\citeyearpar{Lee2023SAME}, to ensure fair comparisons. The source-target ground-truth motion pairs in Mixamo are provided by Adobe, while the character pairings used for evaluation follow these previous methods. Following SAME, we exclude motion clips captured with objects, resulting in 95 motion clips per character instead of the 106 clips in the original evaluation set. The Mixamo evaluation consists of four characters for cross-structural retargeting (cross Mixamo dataset) and one additional character for intra-structural retargeting (intra Mixamo dataset). Our evaluation datasets comprise a diverse set of skeletons with varying joint counts and topological differences.

\subsection{Training Protocol}\label{app:training_protocol}

The training protocol, including hardware and convergence criteria, is summarized in Table~\ref{tab:training_protocol}. Our experiments were conducted on high-performance GPUs to ensure efficient model training.

\begin{table}[ht]
\centering
\caption{Training Protocol Details}
\label{tab:training_protocol}
\begin{tabular}{ll}
\toprule
\textbf{Specification} & \textbf{Details} \\
\midrule
Hardware & NVIDIA Tesla A100 \\
Training Time & Approximately 48 hours \\
\bottomrule
\end{tabular}
\end{table}

%\subsection{Code Availability and Reproducibility}

%The complete implementation, along with pretrained models and detailed instructions for reproducing our results, is publicly available at \url{https://github.com/XXXX/XXXX}. The repository includes guidelines for setting up the environment and executing the training pipeline.

\section{\MakeUppercase{Data Augmentation}}\label{app:data_augment}

In this section, we introduce two complementary data augmentation techniques designed to enrich skeletal animation data by perturbing its geometric properties while preserving the underlying kinematic structure: \emph{skeleton augmentation} and \emph{bone scaling augmentation}. Furthermore, we need \emph{global translation augmentations} to enable the disentanglement of the pose and trajectory latent space. While training on the humanoid characters we only use the three data augmentation mentioned previously. However, when it comes to training on also non-humanoid characters, we also need \emph{rest pose augmentations} to enrich the rest pose variety.

\subsection{Rest Pose Augmentation}\label{app:restpose_augment}

We perturb the skeletal animation’s rest pose by applying random rotations to each joint while maintaining the hierarchical relationships between them. Let \(N\) denote the total number of joints in the skeleton, indexed by \(i = 0, 1, \dots, N-1\). For each joint \(i\), let \(p(i)\) represent the index of its parent, with \(p(0) = -1\) indicating that the root joint has no parent. The original rotations are given by matrices \(R_i \in SO(3)\), as obtained from the animation data.

For augmentation, we independently sample a set of random rotation matrices \(\{Q_i\}_{i=0}^{N-1}\), where each \(Q_i \in SO(3)\) is drawn from a exponential distribution with $\frac{\lambda}{2}\cdot\exp(-\lambda |x| )$ with $\lambda=15$ from the rotation group. These matrices perturb the rest pose without compromising the skeletal hierarchy. The augmented rotation \(R'_i\) for each joint is computed as follows. For the root joint (\(i = 0\)), which has no parent, the augmented rotation is defined by
\[
R'_0 = R_0 \, Q_0^\top,
\]
where \(Q_0^\top\) (the transpose, equivalently the inverse, of \(Q_0\)) counteracts the random offset applied to the root. For a non-root joint (\(i > 0\)) with parent \(p(i)\), the augmented rotation is given by
\[
R'_i = Q_{p(i)} \, R_i \, Q_i^\top.
\]
Here, left-multiplication by \(Q_{p(i)}\) aligns the joint’s coordinate system with the perturbed frame of its parent, while right-multiplication by \(Q_i^\top\) compensates for the joint-specific random rotation. This formulation preserves the relative orientation between parent and child joints despite the applied perturbations.

Each joint is associated with an offset vector \(t_i\) representing its positional displacement from its parent. To maintain consistency after rotating the joints, these offsets are updated as
\[
t'_i = Q_{p(i)} \, t_i,
\]
for every joint \(i\) with \(p(i) \neq -1\).

\subsection{Bone Scaling Augmentation}\label{app:bone_Scaling_aug}

In addition to reorienting the rest pose, we apply a global scaling augmentation to the entire skeletal structure. This strategy uniformly scales the joint offsets, thereby affecting the computed joint positions and the resulting motion dynamics while preserving the overall pose configuration.

Let \(t_i\) denote the original offset vector for joint \(i\). The augmented offset \(\tilde{t}_i\) is obtained by applying a global scaling factor \(s\) to every offset:
\[
\tilde{t}_i = s \cdot t_i, \quad \forall i.
\]
The scaling factor \(s\) is sampled probabilistically as
\[
s =
\begin{cases}
\text{Uniform}(0.5, 1.5), & \text{with probability } 0.25,\\[1mm]
1.0, & \text{with probability } 0.75.
\end{cases}
\]
After scaling the offsets, the new joint positions are recalculated using forward kinematics. Denote by \(p_i\) the original position of joint \(i\) and by \(R_i\) its rotation matrix. The updated joint position \(\tilde{p}_i\) is given by
\[
\tilde{p}_i = f\left(\tilde{t}_i, R_i\right),
\]
where \(f\) is the forward kinematics function that propagates the scaled offsets along the skeletal hierarchy. The same is also applied for the previous joint positions, which is part of the input feature.
Furthermore, the root trajectory \(r\) is updated to \(\tilde{r}\) via
\[
\tilde{r} = s \cdot r.
\]
The joint velocity \(v_i\) is then recomputed based on the difference between the positions in consecutive frames:
\[
v_i = \tilde{p}_i(t) - \tilde{p}_i(t-1).
\]
This recalculation ensures that the dynamic properties of the motion remain consistent with the new, scaled geometry.

\subsection{Skeleton Augmentation}\label{app:skeleton_augmentation}

In order to get a more diverse skeleton dataset, we combine global scaling, local scaling and joint removal augmentations to each skeletal structure used in the cycle consistency step. Let \(t_i\) denote the original offset vector for joint \(i\). 

The global scaling procedure is done in the same way as described in subsection \ref{app:bone_Scaling_aug}. Assume in the following, that \(\tilde{t}_i\) is the global scaling augmented offset, which is obtained by applying a global scaling factor \(s\) to every offset. In order to perform the local scaling augmentation, we first draw a scaling factor $s_{i}$ for each joint in the skeleton. This is sampled independent from other joint scaling factors probabilistically as: 

\[
s_i =
\begin{cases}
\text{Uniform}(0.5, 1.5), & \text{with probability } 0.25,\\[1mm]
1.0, & \text{with probability } 0.75.
\end{cases}
\]

The already globally scaled offset is then scaled for each joint with its respective joint scaling value.

\[
\tilde{t}_i \to s_i \cdot \tilde{t}_i, \quad \forall i.
\]

In the last step, we also further augment the skeleton, by introducing random joint removal. First we need the determine the number of joints, which will be removed. This number $N_\text{to remove}$ is sampled probabilistically as: 

\begin{equation*}
    N_\text{to remove} = \text{Uniform}(0,N_\text{max num}).
\end{equation*}

$N_\text{max num}$ refers to the maximal number of joints, which will be removed and is a hyperparameter. In our case, $N_\text{max num}$ is chosen to be 4. After that, we determine which joints specifically will be removed. For that, we again sample $N_\text{to remove}$ numbers, which are in the range of the number of joints, leaving out zero since this is referred to as the root joint. Each joint which should be removed $J_{i,\text{to remove}}$ will be probabilistically sampled in the following way: 

\begin{equation*}
    J_{i,\text{to remove}} = \text{Uniform}(1,N_\text{joints} - 1).
\end{equation*}

Here we use $N_\text{joints} - 1$, because we start counting from zero on. After we sampled all our $J_{i,\text{to remove}}$, we only keep the unique joint number $\text{Unique}(J_{\text{to remove}})$. 
At the end we need to update the skeleton, which is described by its parents and offsets. Assume in the following, that $C_{i}$ are all the children of joint $i$ and $p_i$ is the parent of joints $i$. Then we update the parent child relation in the following way:

\begin{equation*}
    p_j = p_i \quad  \forall i\in \text{Unique}(J_{\text{to remove}}),\forall j\in C_i.
\end{equation*}

For the offsets, we update it in the following way:

\begin{equation*}
    \tilde{t}_j \to \tilde{t}_j + \tilde{t}_i\quad  \forall i\in \text{Unique}(J_{\text{to remove}}),\forall j\in C_i.
\end{equation*}

After that, all joints in the set $\text{Unique}(J_{\text{to remove}})$ in the parents-child hierarchy as well as for the offsets will be removed. 

\subsection{Global Translation Augmentation}\label{app:global_translation_augmentation}

To further enhance robustness to variations in absolute positioning, we introduce a global translation augmentation that shifts the entire skeletal pose along the horizontal plane. This augmentation simulates variations in subject placement by adding random offsets in the \(x\) and \(z\) coordinates.
Let \(T = (T_x, T_z)\) denote the translation vector applied to the global pose. For each sample, the translation components \(T_x\) and \(T_z\) are independently sampled from a uniform distribution:
\[
T_x \sim \text{Uniform}(a_x, b_x), \quad T_z \sim \text{Uniform}(a_z, b_z),
\]
where \(a_x\), \(b_x\), \(a_z\), and \(b_z\) are hyperparameters defining the translation range.
After sampling \(T\), the translation is applied uniformly to all joint positions. For each joint with original position \(p_i = (x_i, y_i, z_i)\), the updated position \(\tilde{p}_i\) is computed as
\[
\tilde{p}_i = (x_i + T_x,\, y_i,\, z_i + T_z).
\]
This augmentation is also applied to any global trajectories associated with the pose, ensuring that the spatial dynamics remain consistent with the translation.

\subsection{Limitations of the Sparse Representation}

Although the formulation above allows exact recovery of \(q\) from \(\bar{q}\), it has two significant limitations. First, the dimensionality of \(\bar{q}\) grows linearly with the number of joints \(J\), which is suboptimal for deep learning architectures that require fixed-size inputs. Second, due to the sparsity of \(M\) and \(T\), only a subset of the latent dimensions is actively utilized, potentially constraining the expressive capacity of the representation. These challenges motivate our design of a dense, constant-dimensional representation inspired by the operations described above.

\section{\MakeUppercase{Further Results}}

\subsection{Loss Ablation Studies}\label{appendix:loss_ablation}

We ablate each loss individually to verify its contribution to reconstruction and retargeting (Tab. \ref{tab:retargeting_loss_ablation}). \textit{Structural losses:} removing $\mathcal{L}_{\text{child}}$ substantially increases GJP ($1.45 \rightarrow 3.21$ Intra, $1.28 \rightarrow 2.51$ Cross), while removing $\mathcal{L}_{\text{ee}}$ causes smaller but consistent degradations. $\mathcal{L}_{\text{cyc.\ cons.}}$ is essential for cross-structural transfer, increasing Cross GJP from $1.28 \rightarrow 3.25$ when removed, whereas $\mathcal{L}_{\text{adv}}$ provides smaller but consistent improvements. 

removing $\mathcal{L}_{\text{jerk}}$ or $\mathcal{L}_{\text{vel}}$ degrades performance, with $\mathcal{L}_{\text{jerk}}$ notably increasing Cross GJP to $2.07$ due to overly smooth but inaccurate motions.

$\mathcal{L}_{\text{cntct\_vel}}$ has a stronger impact than $\mathcal{L}_{\text{cntct\_pos}}$, particularly on Intra retargeting ($1.45 \rightarrow 2.26$). Removing $\mathcal{L}_{\text{cntct\_pos}}$ slightly improves Intra but harms Cross performance, the more challenging and practically relevant setting, suggesting it mainly benefits cross-topology generalization. $\mathcal{L}_{\text{grnd\_pen}}$ further improves overall retargeting stability and performance across both settings.

We find that our design choices provide at minimum a 11\% intra, and 13\% cross improvement on GJP. However, $\mathcal{L}_\text{cyc. cons.}$ by design only improves GJP on the cross task, and $\mathcal{L}_\text{cntct\_pos}$ increases the GJP error (6\% intra; 3\% cross), however, decreases foot sliding.
\begin{table}[h]
\caption{Ablation Study on the Mixamo dataset. We evaluate the impact of removing key components on Intra- and Cross-retargeting performance.
\textbf{Bold} indicates best; \underline{underline} indicates second best.}
\centering
\begin{tabular}{lcccccc}
\toprule
& \multicolumn{3}{c}{Intra} & \multicolumn{3}{c}{Cross} \\
\cmidrule(lr){2-4} \cmidrule(lr){5-7}
Method & GJP & Jerk & PART & GJP & Jerk & PART \\
\midrule
Ours Full& \underline{1.45} & 0.72 & \textbf{0.03} & \textbf{1.28} & 0.72 & \textbf{0.05}\\
w/o $\mathcal{L}_\text{cyc. cons.}$ & 1.47 & \textbf{0.82} & \textbf{0.03} & 3.25 & \textbf{1.23} & \underline{0.06}\\
w/o $\mathcal{L}_\text{adv}$. & 1.65 & 0.67 & \textbf{0.03} & 1.47 & 0.69 & \textbf{0.05}\\
w/o $\mathcal{L}_\text{child}$ & 3.21 & 0.77 & 0.05 & 2.51 & 0.75 & 0.06\\
w/o $\mathcal{L}_\text{ee}$ & 1.62 & 0.70 & \textbf{0.03} & 1.46 & 0.71 & 0.04\\
w/o $\mathcal{L}_\text{vel}$ & 1.69 & 0.66 & \textbf{0.03} & 1.45 & 0.68 & \textbf{0.05}\\
w/o $\mathcal{L}_\text{jerk}$ & 1.87 & 0.68 & \underline{0.04} & 2.07 & 0.68 & \textbf{0.05}\\
w/o $\mathcal{L}_\text{cntct\_pos}$ & \textbf{1.37} & \underline{0.81} & \underline{0.04} & \underline{1.32} & \underline{0.76} & \textbf{0.05}\\
w/o $\mathcal{L}_\text{cntct\_vel}$ & 2.26 & 0.68 & \underline{0.04} & 1.74 & 0.71 & \textbf{0.05}\\
w/o $\mathcal{L}_\text{grnd\_pen}$ & 1.70 & 0.72 & \underline{0.04} & 1.51 & 0.72 & \textbf{0.05}\\
\bottomrule
\end{tabular}
\label{tab:retargeting_loss_ablation}
\end{table}

\clearpage
\subsection{Trajectory Error Accumulation }\label{appendix:traj_err_accumulation}

\begin{figure*}[h]
  \centering
  \includegraphics[width=0.9\linewidth]{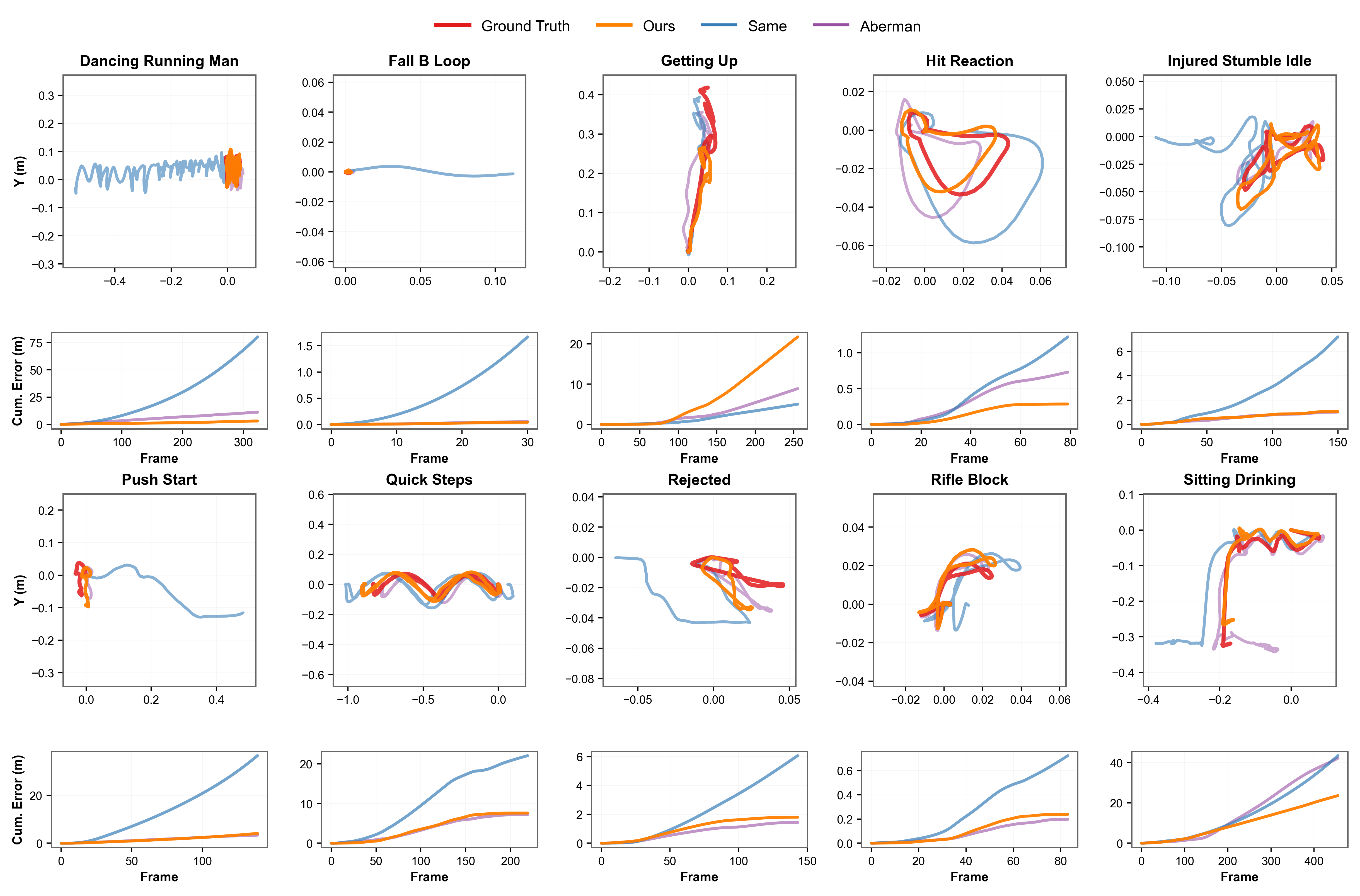}
  \caption{Error accumulation when using the delta representation for trajectories:. In contrast, predicting the global position (Ours, \citet{Aberman2020}) does not result in as much error accumulation (SAME). It shows the global root trajectory projected on the xz plane (top) and the xz root trajectory error with increasing number of frames (bottom) for three different motions. The global root trajectory prediction method used by us and \citet{Aberman2020} does not suffer from error accumulation, unlike SAME who predicts only the difference between two frames.}
  \label{fig:trajectory_analysis}
\end{figure*}

\subsection{Skeleton Invariance}\label{appendix:skeleton:topology_invariance}
% \begin{figure*}[h]
%   \centering
%   \includegraphics[width=0.9\linewidth]{figures/latent_trajectories.png}
%   \caption{Skeleton invariance latent spaces of the model: 2D Principle Component Analysis projected latent space of the model for different skeletons performing semantically identical motions for different motions. The PCA space is shared across all sequences.}
%   \label{fig:skeleton_invariance}
% \end{figure*}
\clearpage
\begin{figure*}[h]
  \centering
  \includegraphics[width=0.9\linewidth]{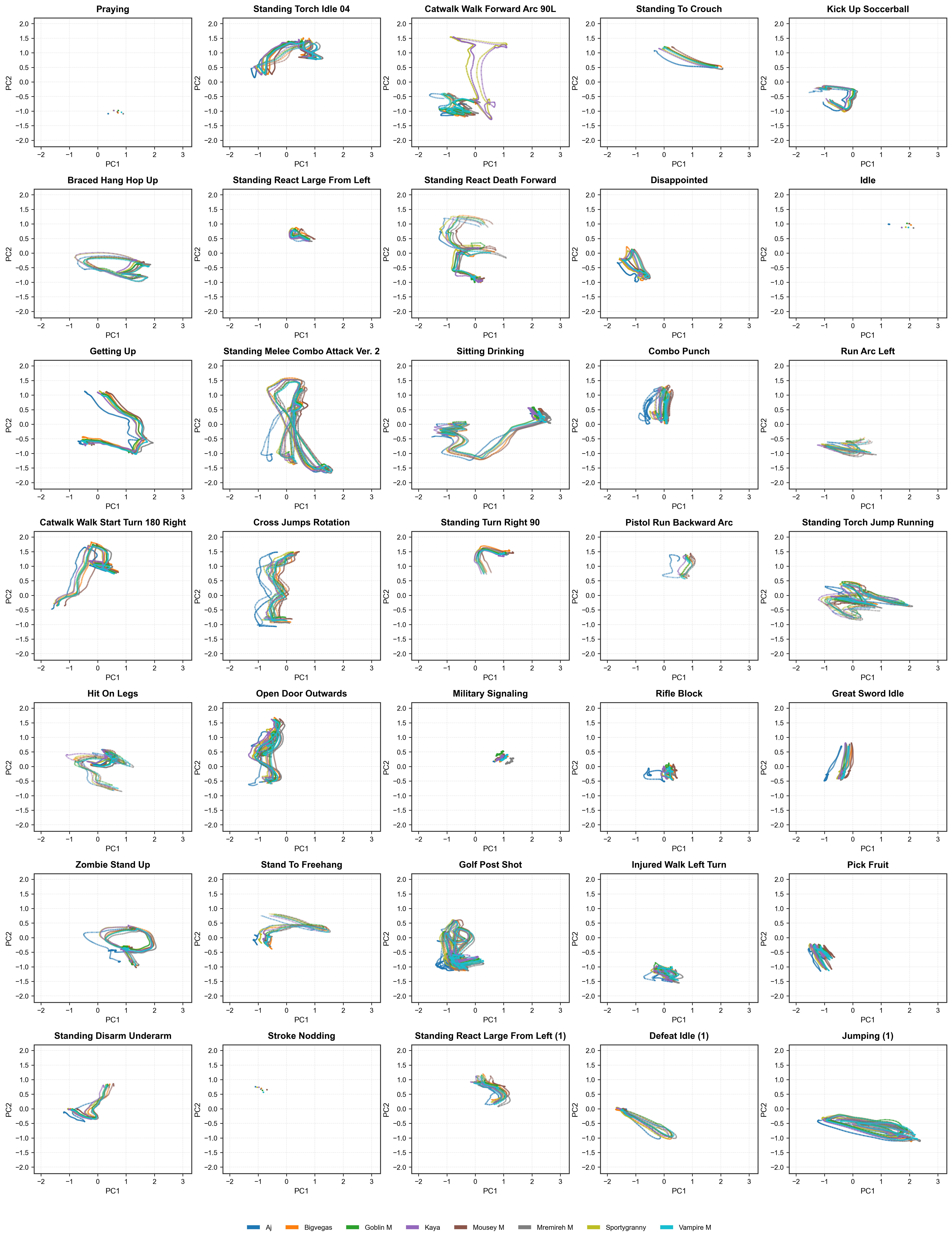}
  \caption{Additional skeleton invariance latent spaces of the model: 2D Principal Component Analysis projected latent space of the model for different skeletons performing semantically identical motions for different motions. The PCA space is shared across all sequences.}
  \label{fig:latent_trajectories_extra}
\end{figure*}
\clearpage
% \subsection{Translation Invariance}\label{appendix:translation_invariance}

% \begin{figure*}[h]
%   \centering
%   \includegraphics[width=1\linewidth]{figures/translation_invariance_combined.png}
%   \caption{Translation invariance of our pose latent space. Left shows distributions of cosine similarities between latents of untranslated and translated motions. SAME is invariant by definition, so all the values are 1. Right shows the mean cosine similarities across translations in the x-, z- and xz-axes.}
%   \label{fig:translation_invariance}
% \end{figure*}

\subsection{Retargeting}\label{appendix:retargeting}
% \begin{figure*}[h]
%   \centering
%   \includegraphics[width=\linewidth]{figures/fig-retarget-comparison.png}
%   \caption{Retargeting across diverse skeletons: The green skeletons are the source motions, red are ground truth retargets from the Mixamo dataset,  the orange are retargets from our method (SKiP), blue are from SAME and purple from \citet{Aberman2020}.}
%   \label{fig:retarget_paper_comparison}
% \end{figure*}

\begin{figure*}[h]
  \centering
  \includegraphics[width=\linewidth]{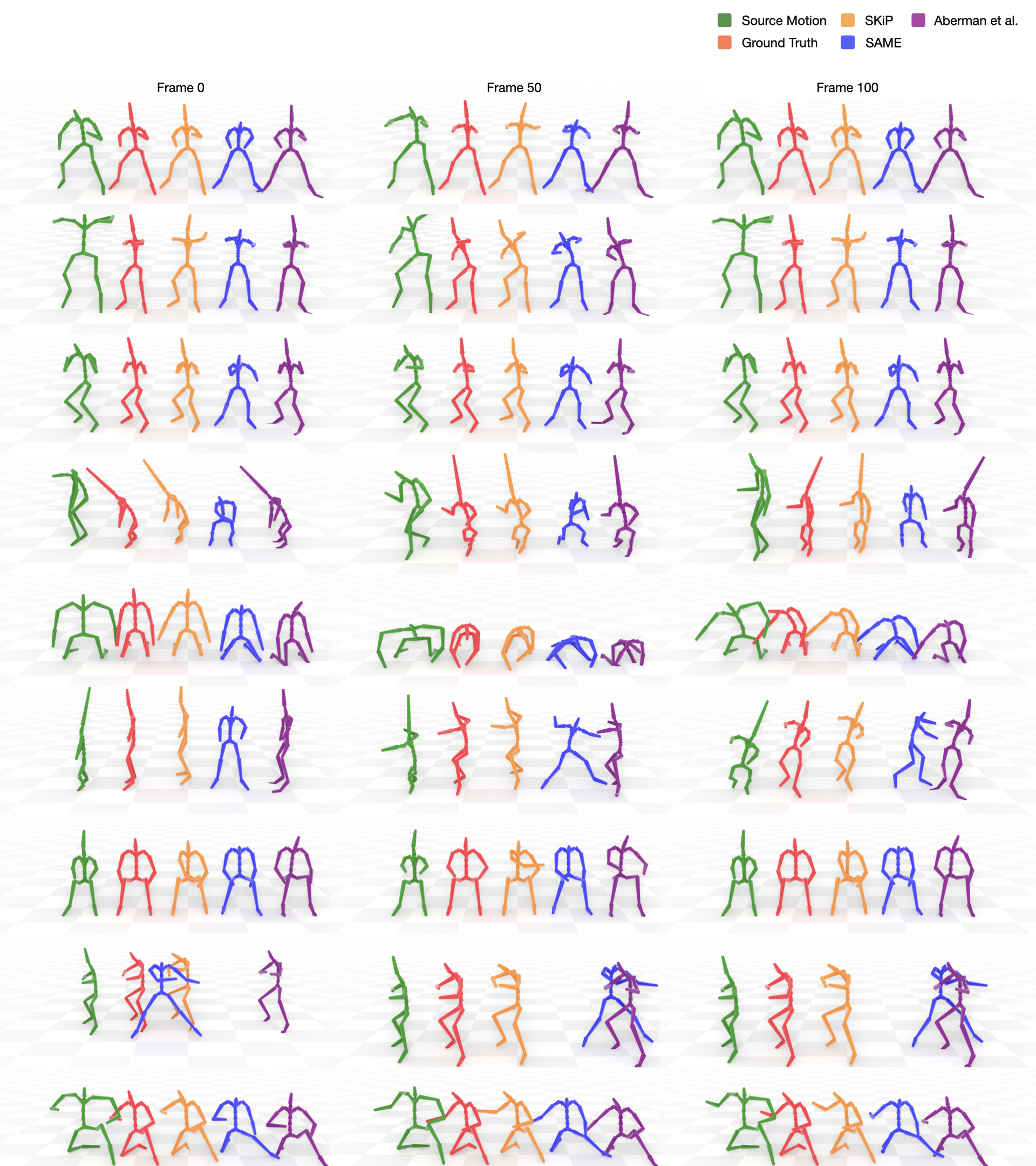}
  \caption{Additional examples of retargeting across diverse skeletons: The green skeletons are the source motions, red are ground truth retargets from the Mixamo dataset,  the orange are retargets from our method, blue are from SAME and purple from \citet{Aberman2020}. Each row represents the same motion, while each column frames 0, 50 and 100 from that clip.}
  \label{fig:retarget_paper_extra}
\end{figure*}
\end{document}